\documentclass[10pt,journal,compsoc]{IEEEtran}

\ifCLASSOPTIONcompsoc
  \usepackage[nocompress]{cite}
\else
  \usepackage{cite}
\fi
\usepackage{graphicx}
\usepackage{amsmath,amssymb} 
\usepackage{dsfont}
\usepackage{color}
\usepackage{tikz}
\usepackage{multirow}
\usepackage{xspace}
\usepackage{bbm}
\usepackage{fixltx2e}
\usepackage{contour}
\usepackage{colortbl}
\usepackage{array}
\usepackage[]{units}
\usepackage{xfrac}
\usepackage{flushend}

\usepackage[nomessages]{fp}

\usepackage[pagebackref=true,breaklinks=true,colorlinks,bookmarks=false]{hyperref} 

\DeclareMathOperator*{\argmin}{\arg\!\min}
\DeclareMathOperator*{\argmax}{\arg\!\max}

\usepackage{pgfplots}
\usepackage{pgfplotstable}
\usepackage{pgf,tikz}

\usepgfplotslibrary{external}
\newcommand{\extfig}[2]{\tikzsetnextfilename{fig/extern/#1}{#2}}
\newcommand{\extdata}[1]{\input{#1}}

\newcommand{\leg}[1]{\addlegendentry{#1}}

\begin{document}
%
\bstctlcite{IEEEexample:BSTcontrol}
%
\title{Fine-tuning CNN Image Retrieval\\
with No Human Annotation}
%
%
%
%

\newcommand{\namespace}{\hspace{5mm}}
\author{Filip Radenovi{\'c} \namespace Giorgos Tolias \namespace Ond{\v r}ej Chum \protect\\ 
\thanks{
\textbf{Affiliation:} Visual Recognition Group, 
Department of Cybernetics, 
Faculty of Electrical Engineering, 
Czech Technical University in Prague \vspace{0.8mm} \protect\\
\textbf{E-mail:} \{filip.radenovic,giorgos.tolias,chum\}@cmp.felk.cvut.cz \protect\\
\textbf{Data, networks, and code:} \href{http://cmp.felk.cvut.cz/cnnimageretrieval}{cmp.felk.cvut.cz/cnnimageretrieval}
}
}

\def\ie{\emph{i.e.}\xspace}
\def\eg{\emph{e.g.}\xspace}
\def\wrt{\emph{w.r.t.}\xspace}
\def\etal{\emph{et al.}\xspace}
\def\etc{\emph{etc.}\xspace}

\newcommand{\gem}{GeM\xspace}

\definecolor{darkred}{rgb}{0.72,0.11,0.11}
\definecolor{darkblue}{rgb}{0.10,0.14,0.79}
\definecolor{redd}{rgb}{0.85,0.26,0.08}
\definecolor{greenn}{rgb}{0.30,0.69,0.31}

\newcommand{\real}{\mathbb{R}}
\newcommand{\realnn}{{\mathbb{R}^{+}_{0}}}
\newcommand{\nat}{\mathbb{N}}
\newcommand{\natzero}{{\mathbb{N}_{0}}}

\newcommand{\loss}{\mathcal{L}}

\newcommand{\cX}{\mathcal{X}}
\newcommand{\cI}{\mathcal{I}}
\newcommand{\cP}{\mathcal{P}}
\newcommand{\cE}{\mathcal{E}}
\newcommand{\cN}{\mathcal{N}}
\newcommand{\cM}{\mathcal{M}}

\newcommand{\bG}{\mathbb{G}}
\newcommand{\cG}{\boldsymbol{\mathcal{G}}}

\newcommand{\vf}{\mathbf{f}}
\newcommand{\f}{\mathrm{f}}
\newcommand{\mac}{\bar{\vf}}

\def\l2{$\ell_2$}

\xspaceaddexceptions{+}
\def\cpl2{L\textsubscript{w}\xspace}
\def\pcawhiten{PCA\textsubscript{w}\xspace}

\def\cropI{$\texttt{Crop}_\cI$\xspace}
\def\cropA{$\texttt{Crop}_\cX$\xspace}

\renewcommand{\paragraph}[1]{{\medskip \noindent \bf #1}}
\newcommand{\pari}[1]{{\medskip \noindent \it #1}}
\newcommand{\equ}[1]{Equation~(\ref{#1})\xspace}

\newcommand{\gio}[1]{{\color{purple}{#1}}}
\newcommand{\fr}[1]{{\color{blue}{#1}}}
\newcommand{\alert}[1]{#1}

\renewcommand{\b}[1]{\textbf{#1}}
\newcommand{\w}[1]{\color{blue}{#1}}
\newcommand{\ww}[1]{\textbf{\color{blue}{#1}}}

\newcommand{\ob}[1]{\textbf{#1}}
\newcommand{\nb}[1]{{\color{black}{\contourlength{0.3pt}\contour{darkred}{#1}}}}
\renewcommand{\sb}[1]{{\color{black}{\contour{darkred}{#1}}}}
\newcommand{\bo}{\cellcolor{gray!15}}

\def\sssp{\hspace{1pt}}
\def\ssp{\hspace{3pt}}
\def\msp{\hspace{5pt}}
\def\bsp{\hspace{8pt}}
\def\nsp{\hspace{-2pt}}

\IEEEtitleabstractindextext{%
\begin{abstract}
Image descriptors based on activations of Convolutional Neural Networks (CNNs) have become dominant in image retrieval due to their discriminative power, compactness of representation, and search efficiency. 
Training of CNNs, either from scratch or fine-tuning, requires a large amount of annotated data, where a high quality of annotation is often crucial.
In this work, we propose to fine-tune CNNs for image retrieval on a large collection of unordered images in a fully automated manner.  
Reconstructed 3D models obtained by the state-of-the-art retrieval and structure-from-motion methods guide the selection of the training data.
We show that both hard-positive and hard-negative examples, selected by exploiting the geometry and the camera positions available from the 3D models, enhance the performance of particular-object retrieval.
CNN descriptor whitening discriminatively learned from the same training data outperforms commonly used PCA whitening.
We propose a novel trainable Generalized-Mean (\gem) pooling layer that generalizes max and average pooling and show that it boosts retrieval performance.
Applying the proposed method to the VGG network achieves state-of-the-art performance on the standard benchmarks: Oxford Buildings, Paris, and Holidays datasets.
\end{abstract}
}
\maketitle

\IEEEdisplaynontitleabstractindextext

%
\IEEEpeerreviewmaketitle

%
\begin{figure*}[t]
\centering

\includegraphics[width=0.95\linewidth]{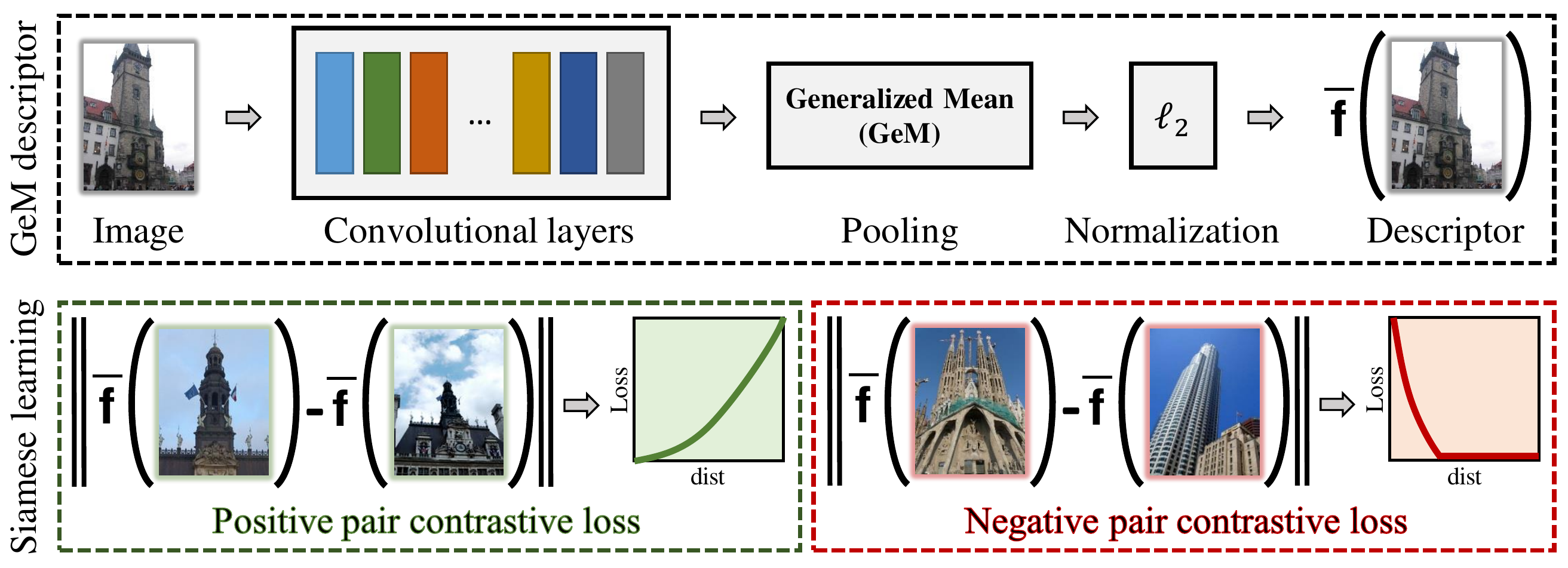}
\vspace{-5pt}
\caption{The architecture of our network with the contrastive loss used at training time. A single vector $\mac$ is extracted to represent an image.
\label{fig:network}
\vspace{-10pt}
}
\end{figure*}
%
\IEEEraisesectionheading{\section{Introduction}\label{sec:introduction}}
%
\IEEEPARstart{I}{n} instance image retrieval an image of a particular object, depicted in a query, is sought in a large, unordered collection of images.  Convolutional neural networks (CNNs) have recently provided an attractive solution to this problem. 
In addition to leaving a small memory footprint, the CNN-based approaches achieve high accuracy.
Neural networks have attracted a lot of attention after the success of Krizhevsky \etal~\cite{KSH12} in the image-classification task.
Their success is mainly due to the use of very large annotated datasets, \eg ImageNet~\cite{RDSK+15}. 
The acquisition of the training data is a costly process of manual annotation, often prone to errors.
Networks trained for image classification have shown strong adaptation abilities~\cite{ARSM+14}.
Specifically, using activations of CNNs, which were trained for the task of classification, as off-the-shelf image descriptors~\cite{DJVO+13,RASC14} and adapting them for a number of tasks~\cite{GDDM14,IMKG+14,GWGL14} have shown acceptable results. 
In particular, for image retrieval, a number of approaches directly use the network activations as image features and successfully perform image search~\cite{GWGL14,RSMC14,BL15,KMO15,TSJ16}.

\emph{Fine-tuning} of the network, \ie initialization by a pre-trained classification network and then training for a different task, is an alternative to a direct application of a pre-trained network.
Fine-tuning significantly improves the adaptation ability~\cite{ZDGD14,OBLS14}; however, further annotation of training data is required. 
The first fine-tuning approach for image retrieval is proposed by Babenko \etal~\cite{BSCL14}, in which a significant amount of manual effort was required to collect images and label them as specific building classes.
They improved retrieval accuracy; however, their formulation is much closer to classification than to the desired properties of instance retrieval. 
In another approach, Arandjelovic~\etal~\cite{AGTPS15} perform fine-tuning guided by geo-tagged image databases and, similar to our work, they directly optimize the similarity measure to be used in the final task by selecting \emph{matching} and \emph{non-matching} pairs to perform the training.

In contrast to previous methods of training-data acquisition for image search, we dispense with the need for manually annotated data or any assumptions on the training dataset. We achieve this by exploiting the geometry and the camera positions from 3D models reconstructed automatically by a structure-from-motion (SfM) pipeline.
The state-of-the-art retrieval-SfM pipeline~\cite{SRCF15} takes an unordered image collection as input and attempts to build all possible 3D models. To make the process efficient, fast image clustering is employed.
A number of image clustering methods based on local features have been introduced~\cite{CM10a,WL13,PSZ11}. 
Due to spatial verification, the \emph{clusters} discovered by these methods are reliable. 
In fact, the methods provide not only clusters, but also a matching graph or sub-graph on the cluster images. 
The SfM filters out virtually all mismatched images and provides image-to-model matches and camera positions for all matched images in the cluster. The whole process, from unordered collection of images to detailed 3D reconstructions, is fully automatic.
Finally, the 3D models guide the selection of matching and non-matching pairs.
We propose to exploit the training data acquired by the same procedure in the descriptor post-processing stage to learn a discriminative whitening.

An additional contribution of this work lies in the introduction of a novel pooling layer after the convolutional layers. Previously, a number of approaches have been used. These range from fully-connected layers~\cite{BSCL14,GWGL14}, to different global-pooling layers, \eg max pooling~\cite{RSMC14}, average pooling~\cite{BL15}, hybrid pooling~\cite{MK15}, weighted average pooling~\cite{KMO15}, and regional pooling~\cite{TSJ16}.
We propose a pooling layer based on a generalized-mean that has learnable parameters, either one global or one per output dimension.
Both max and average pooling are its special cases.
Our experiments show that it offers a significant performance boost over standard non-trainable pooling layers.
Our architecture is shown in Figure~\ref{fig:network}.

To summarize, we address the unsupervised fine-tuning of CNNs for image retrieval. 
In particular, we make the following contributions:
(1)~We exploit SfM information and enforce, not only hard non-matching (\emph{negative}), but also hard-matching (\emph{positive}) examples for CNN training. 
This is shown to enhance the derived image representation. 
We show that compared to previous supervised approaches, the variability in the training data from 3D reconstructions delivers superior performance in the image-retrieval task. 
(2)~We show that the whitening traditionally performed on short representations~\cite{JC12} is, in some cases, unstable. We propose to learn the whitening through the same training data. Its effect is complementary to fine-tuning and it further boosts performance. 
Also, performing whitening as a post-processing step is better and much faster to train compared to learning it end-to-end.
(3)~We propose a trainable pooling layer that generalizes existing popular pooling schemes for CNNs. 
It significantly improves the retrieval performance while preserving the same descriptor dimensionality.
(4)~In addition, we propose a novel $\alpha$-weighted query expansion that is more robust compared to the standard average query expansion technique widely used for compact image representations.
(5)~Finally, we set a new state-of-the-art result for Oxford Buildings, Paris, and Holidays datasets by re-training the commonly used CNN architectures, such as AlexNet~\cite{KSH12}, VGG~\cite{SZ14}, and ResNet~\cite{HZRS16}.

This manuscript is an extension of our previous work~\cite{RTC16}. 
We additionally propose a novel pooling layer (Section~\ref{sec:gemp}), a novel multi-scale image representation (Section~\ref{sec:exp_test}), and a novel query expansion method (Section~\ref{sec:exp_results}). 
Each one of the newly proposed methods boosts image-retrieval performance, and is accompanied by experiments that give useful insights. 
In addition, we provide an extended related work discussion including the different pooling procedures used in prior CNN work and descriptor whitening. 
Finally, we compare our approach to the concurrent related work of Gordo \etal~\cite{GARL16,GARL16a}.
They significantly improve the retrieval performance through end-to-end learning which incorporates building-specific region proposals. 
In contrast to their work, we focus on the importance of hard-training data examples, and employ a much simpler but equally powerful pooling layer.

The rest of the paper is organized as follows. 
Related work is discussed in Section~\ref{sec:related}, our network architecture, learning procedure, and search process is presented in Section~\ref{sec:network}, and our proposed automatic acquisition of the training data is described in Section~\ref{sec:dataset}.
Finally, in Section~\ref{sec:experiments} we perform an extensive quantitative and qualitative evaluation of all proposed novelties with different CNN architectures and compare to the state of the art.
\section{Related work}
\label{sec:related}

The CNN-based representation is an appealing solution for image retrieval and in particular for compact image representations.
Previous compact descriptors are typically constructed by an aggregation of local features, where representatives are  Fisher vectors~\cite{PLSP10}, VLAD~\cite{JPDSPS11} and alternatives~\cite{RJC15,AZ13,TFJ14}.
Impressively, in this work we show that CNNs dominate the image search task by outperforming state-of-the-art methods that have reached a higher level of maturity by incorporating large visual codebooks~\cite{PCISZ07,AK12}, spatial verification~\cite{PCISZ07,SLBW14} and query expansion~\cite{CMPM11,DGBQG11,TJ14}.

In this work, instance retrieval is cast as a metric learning problem, \ie, an image embedding is learned so that the Euclidean distance captures the similarity well.
Typical architectures for metric learning, such as the two-branch siamese~\cite{CHL05,HCL06,HLT14} or triplet networks~\cite{WSLT+14,SKP15,HA15} employ \emph{matching} and \emph{non-matching} pairs to perform the training and better suit to this task.
Here, the problem of annotations is even more pronounced, \ie, for classification one needs only object category label, while for particular objects the labels have to be per image pair.
Two images from the same object category could potentially be completely different, \eg, different viewpoints of the building or different buildings.
We solve this problem in a fully automated manner, without any human intervention, by starting from a large unordered image collection.

In the following text we discuss the related work for our main contributions, \ie, the training data collection, the pooling approach to construct a global image descriptor, and the descriptor whitening.

\subsection{Training data}
A variety of previous methods apply CNN activations on the task of image retrieval~\cite{GWGL14,RSMC14,BL15,KMO15,TSJ16,ZZWWT16}.
The achieved accuracy on retrieval is evidence for the generalization properties of CNNs. 
The employed networks are trained for image classification using ImageNet dataset~\cite{RDSK+15} by minimizing classification error.
Babenko \etal~\cite{BSCL14} go one step further and re-train such networks with a dataset that is closer to the target task.
They perform training with object classes that correspond to particular landmarks/buildings. 
Performance is improved on standard retrieval benchmarks.
Despite the achievement, still, the final metric and the utilized layers are different to the ones actually optimized during learning.

Constructing such training datasets requires manual effort. 
In recent work, geo-tagged datasets with timestamps offer the ground for weakly-supervised fine-tuning of a triplet network~\cite{AGTPS15}. 
Two images taken far from each other can be easily considered as non-matching, while matching examples are picked by the most similar nearby images. 
In the latter case, similarity is defined by the current representation of the CNN.
This is the first approach that performs end-to-end fine-tuning for image retrieval and, in particular, for the  geo-localization task.
The used training images are now more relevant to the final task.
We differentiate by discovering matching and non-matching image pairs in an unsupervised way.
Moreover, we derive matching examples based on 3D reconstructions which allows for harder \mbox{examples}.

Even though hard-negative mining is a standard process~\cite{GDDM14,AGTPS15}, this is not the case with hard-positive examples.
Mining of hard positive examples have been exploited in the work Simo-Serra \etal~\cite{STFKM14}, where patch-level examples were extracted though the guidance from a 3D reconstruction.
Hard-positive pairs have to be sampled carefully. Extremely hard positive examples (such as minimal overlap between images or extreme scale change) do not allow to generalize and lead to over-fitting.

A concurrent work to ours also uses local features and geometric verification to select positive examples~\cite{GARL16}.
In contrast to our fully unsupervised method,  they start from a landmarks dataset, which had to be manually cleaned, and the landmark labels of the dataset, rather than the geometry, were used to avoid exhaustive evaluation. The same training dataset is used by Noh \etal~\cite{NAS+17} to learn global image descriptors using a saliency mask. However, during test time the CNN activations are seen as local descriptors, indexed independently, and used for a subsequent spatial-verification stage. Such approach boosts accuracy compared to global descriptors, but at the cost of much higher complexity.

\vspace{2mm}
\subsection{Pooling method}
Early approaches to applying CNNs for image retrieval included methods that set the fully-connected layer activations to be the global image descriptors~\cite{BSCL14,GWGL14}. 
The work by Razavian \etal~\cite{RSMC14} moves the focus to the activations of convolutional layers followed by a global-pooling operation. 
A compact image representation is constructed in this fashion with dimensionality equivalent to the number of feature maps of the corresponding convolutional layer. 
In particular, they propose to use max pooling, which is later approximated with integral max pooling~\cite{TSJ16}.

Sum pooling was initially proposed by Babenko and Lempitsky~\cite{BL15}, which was shown to perform well especially due to the subsequent descriptor whitening. 
One step further is the weighted sum pooling of Kalantidis \etal~\cite{KMO15}, which can also be seen as a way to perform transfer learning. 
Popular encodings such as BoW, VLAD, and Fisher vectors are adapted in the context of CNN activations in the work of Mohedano \etal~\cite{MMOS+16}, Arandjelovic \etal~\cite{AGTPS15}, and Ong \etal~\cite{OHB16}, respectively.
Sum pooling is employed once an appropriate embedding is performed beforehand.

A hybrid scheme is the R-MAC method~\cite{TSJ16}, which performs max pooling over regions and finally sum pooling of the regional descriptors.
Mixed pooling is proposed globally for retrieval~\cite{MK15} and the standard local pooling is used for object recognition~\cite{LGT16}. 
It is a linear combination of max and sum pooling. A generalization scheme similar to ours is proposed in the work of Cohen \etal~\cite{CSS16} but in a different context. They replace the standard local max pooling with the generalized one. Finally, generalized mean is used by Mor{\`e}re \etal~\cite{MLV+17} to pool the similarity values under multiple transformations.

\vspace{2mm}
\subsection{Descriptor whitening}
Whitening the data representation is known to be very essential for image retrieval since the work of J\'egou and Chum~\cite{JC12}.
Their interpretation lies on jointly down-weighting co-occurrences and, thus, handling the problem of over-counting. 
The benefit of whitening is further emphasized in the case of CNN-based descriptors~\cite{RASC14,BL15,TSJ16}.
Whitening is commonly learned from a generative model in an unsupervised way by PCA on an independent dataset.

We propose to learn the whitening transform in a discriminative manner, using the same acquisition procedure of the training data from 3D models. A similar approach has been used to whiten local-feature descriptors by Mikolajczyk and Matas~\cite{MM07}.

In constrast, Gordo \etal~\cite{GARL16} learn the whitening in the CNN in an end-to-end manner.
In our experiments we found this choice to be at most as good as the descriptor post-processing and less efficient due to slower convergence of the learning.

%
\begin{figure*}[t]
\centering

\def\queryone{11}
\def\dbimageone{3885}
\def\querytwo{42}
\def\dbimagetwo{211}
\renewcommand{\arraystretch}{1.2}
\resizebox{.95\textwidth}{!}{
\setlength{\tabcolsep}{0pt}
\hspace{-15pt}
\begin{tabular}{c@{\hspace{10pt}}c@{\hspace{15pt}}c}
\raisebox{10pt}{\includegraphics[height=78pt]{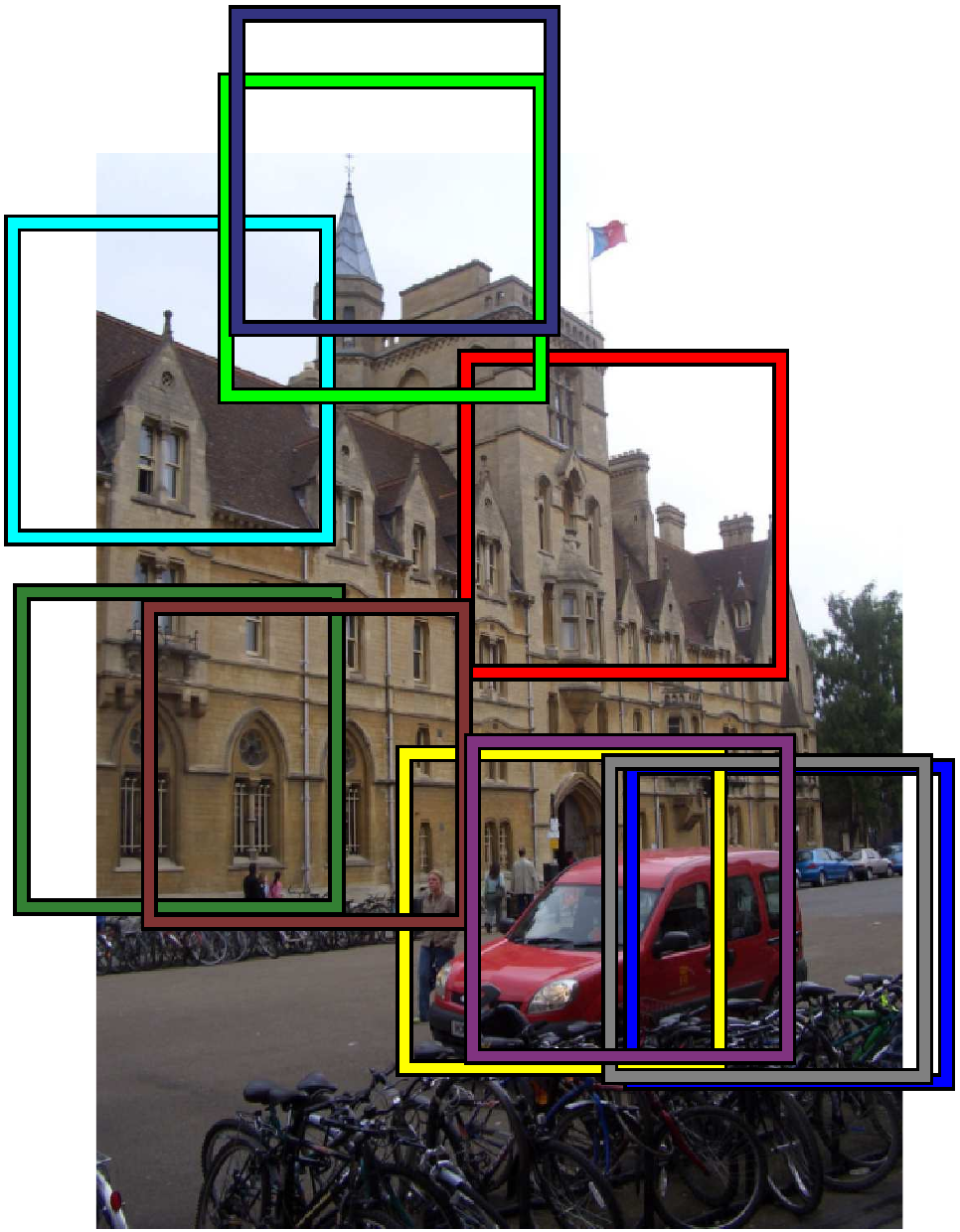} }  &
\includegraphics[height=86pt]{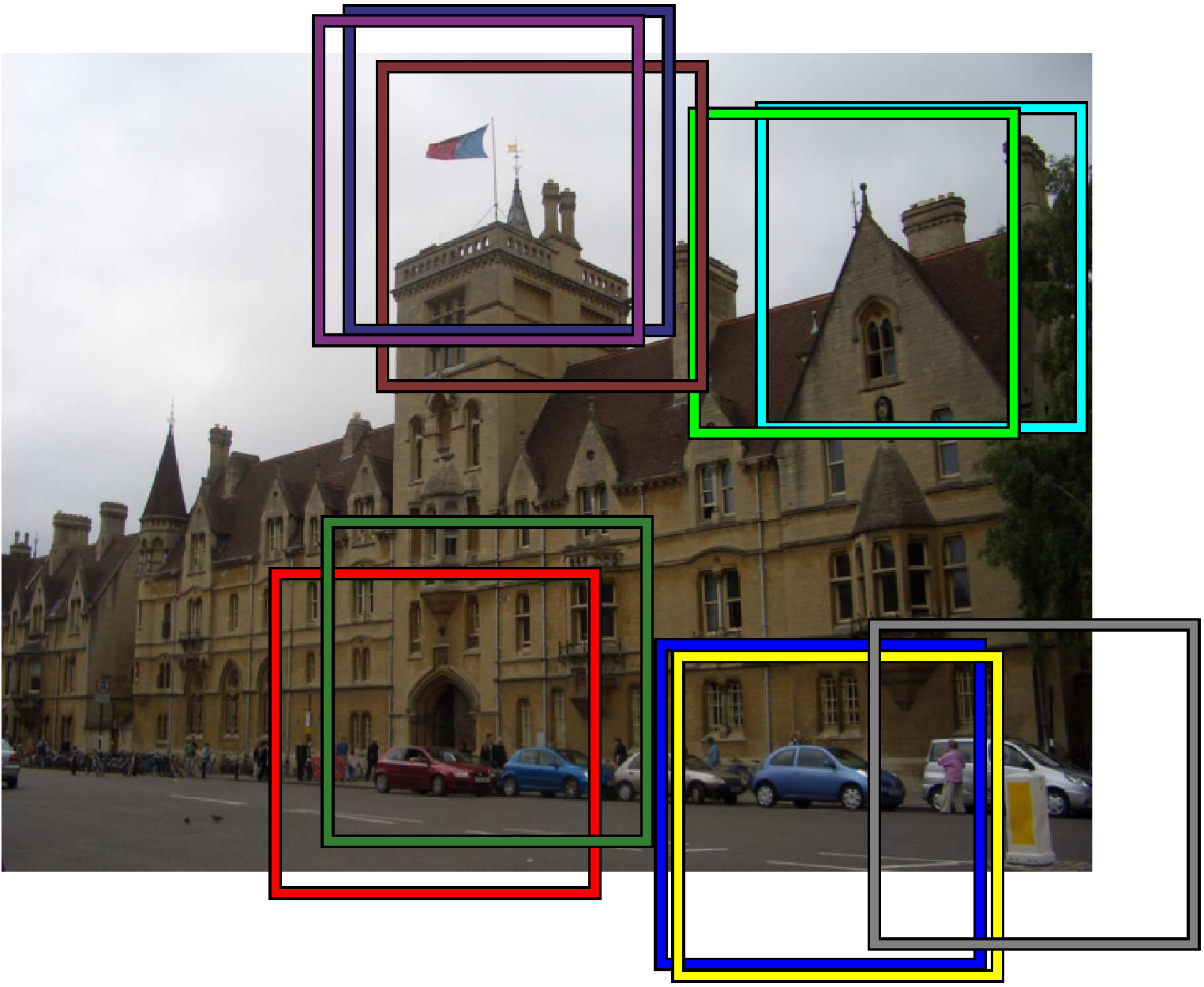}  &
\raisebox{30pt}{
	\begin{tabular}{c}
		\multicolumn{1}{c}{VGG off-the-shelf} \\	
		\foreach \patch in {1,2,3,4,5,6,7,8,9,10}  {
		\includegraphics[height=28pt]{fig/correspondences/q\queryone_db\dbimageone_net0_p\patch_1.png}
		\hspace{-4pt} 
		}\\
		\foreach \patch in {1,2,3,4,5,6,7,8,9,10}  {
		\includegraphics[height=28pt]{fig/correspondences/q\queryone_db\dbimageone_net0_p\patch_2.png} 
		\hspace{-4pt} 
		}\\
	\end{tabular}
}
\\
\includegraphics[height=72pt]{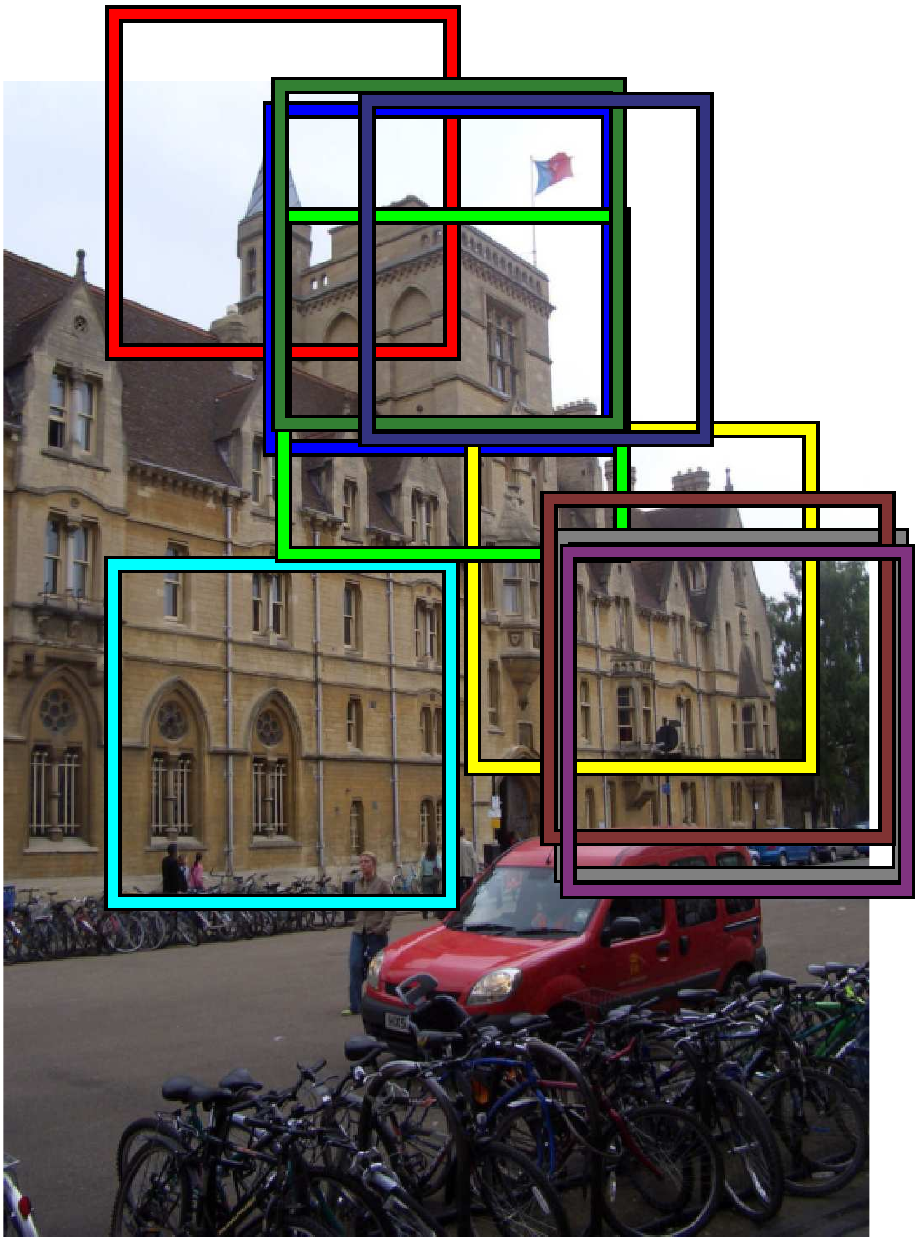} & 
\includegraphics[height=69pt]{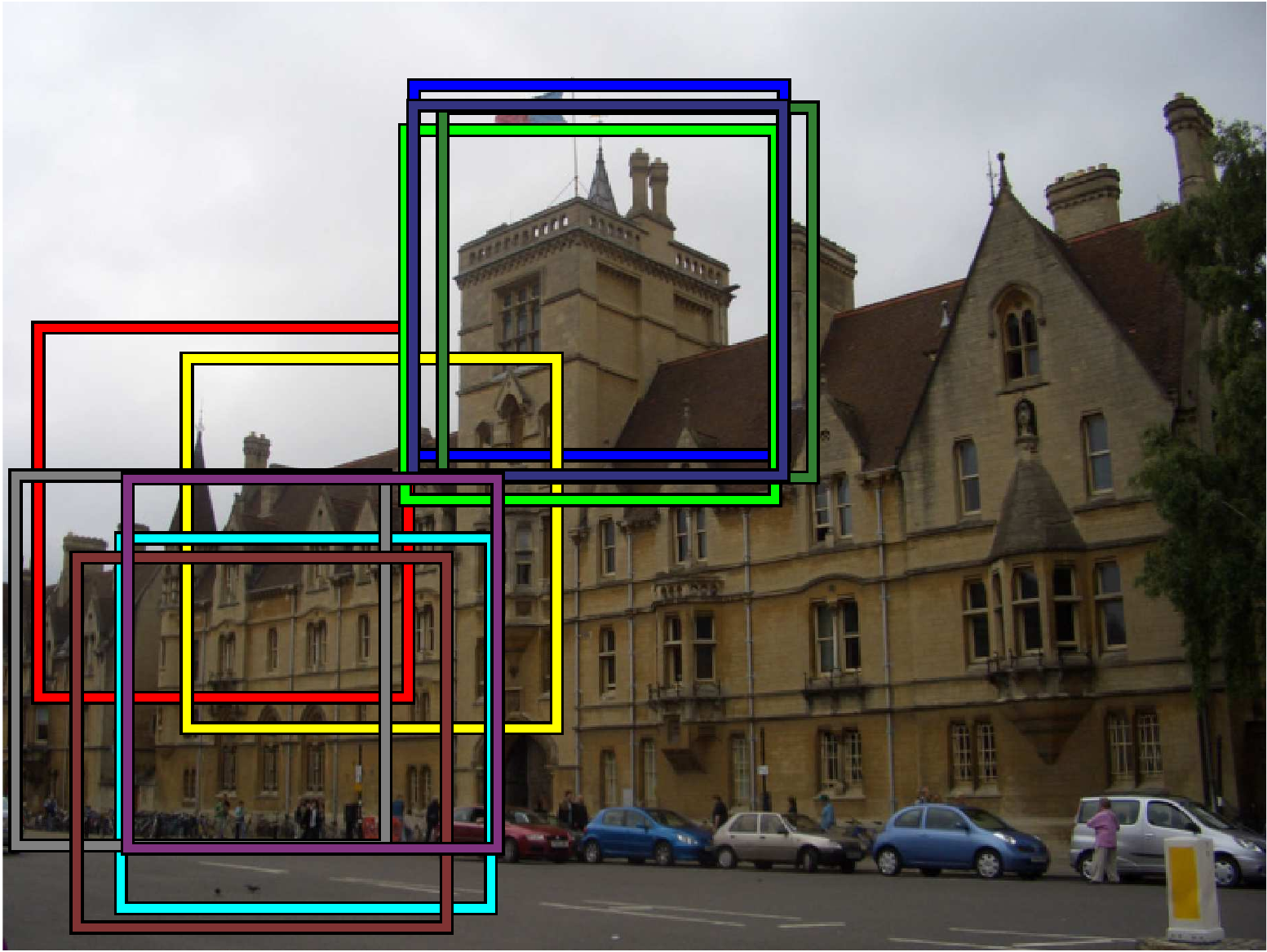} ~~~&
\raisebox{30pt}{
	\begin{tabular}{c}
		\multicolumn{1}{c}{VGG ours} \\	
		\foreach \patch in {1,2,3,4,5,6,7,8,9,10}  {
		\includegraphics[height=28pt]{fig/correspondences/q\queryone_db\dbimageone_net1_p\patch_1.png}
		\hspace{-4pt} 
		}\\
		\foreach \patch in {1,2,3,4,5,6,7,8,9,10}  {
		\includegraphics[height=28pt]{fig/correspondences/q\queryone_db\dbimageone_net1_p\patch_2.png} 
		\hspace{-4pt} 
		}\\
	\end{tabular}
}
\\
\includegraphics[height=84pt]{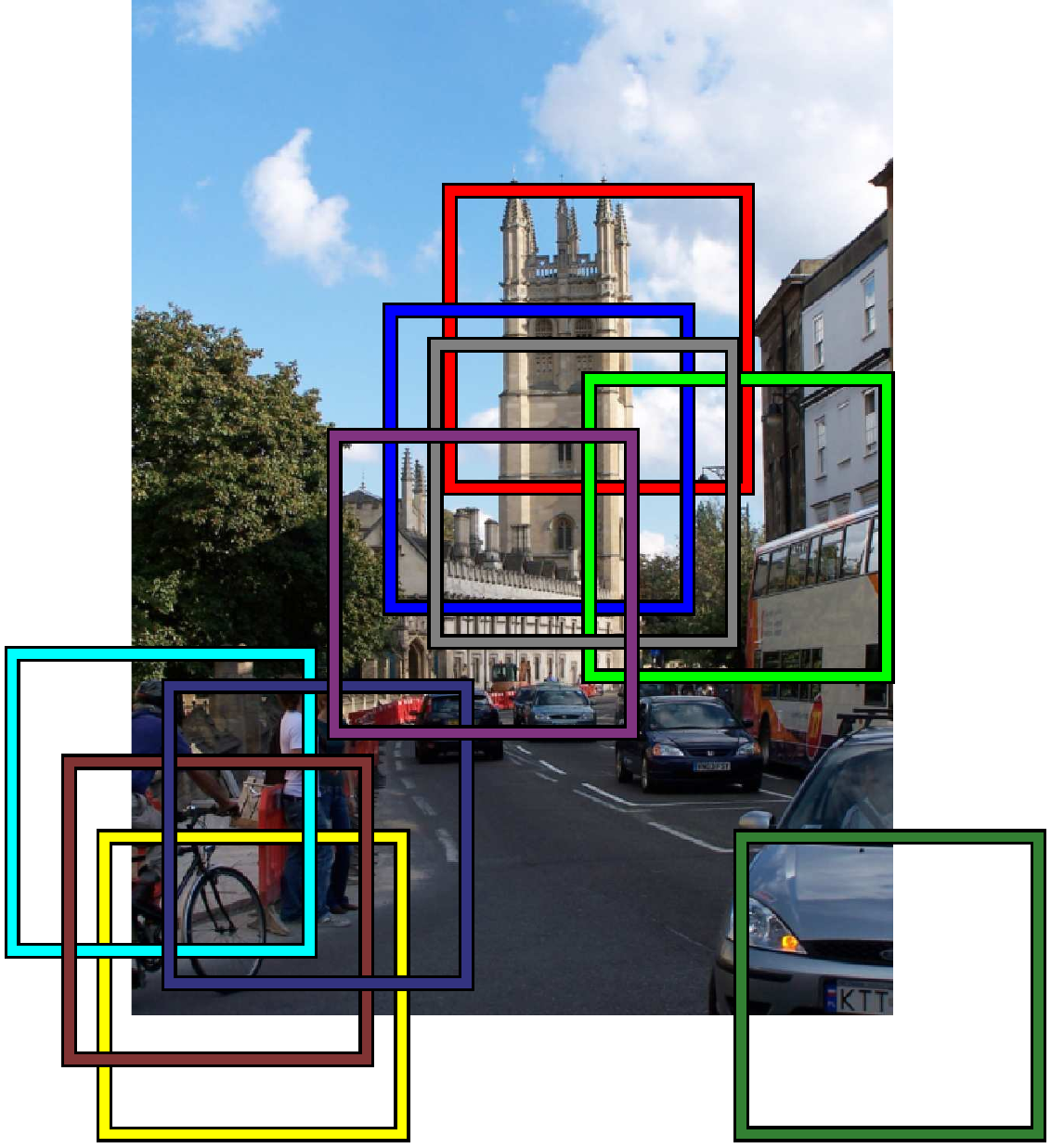} &
\raisebox{8pt}{\includegraphics[height=84pt]{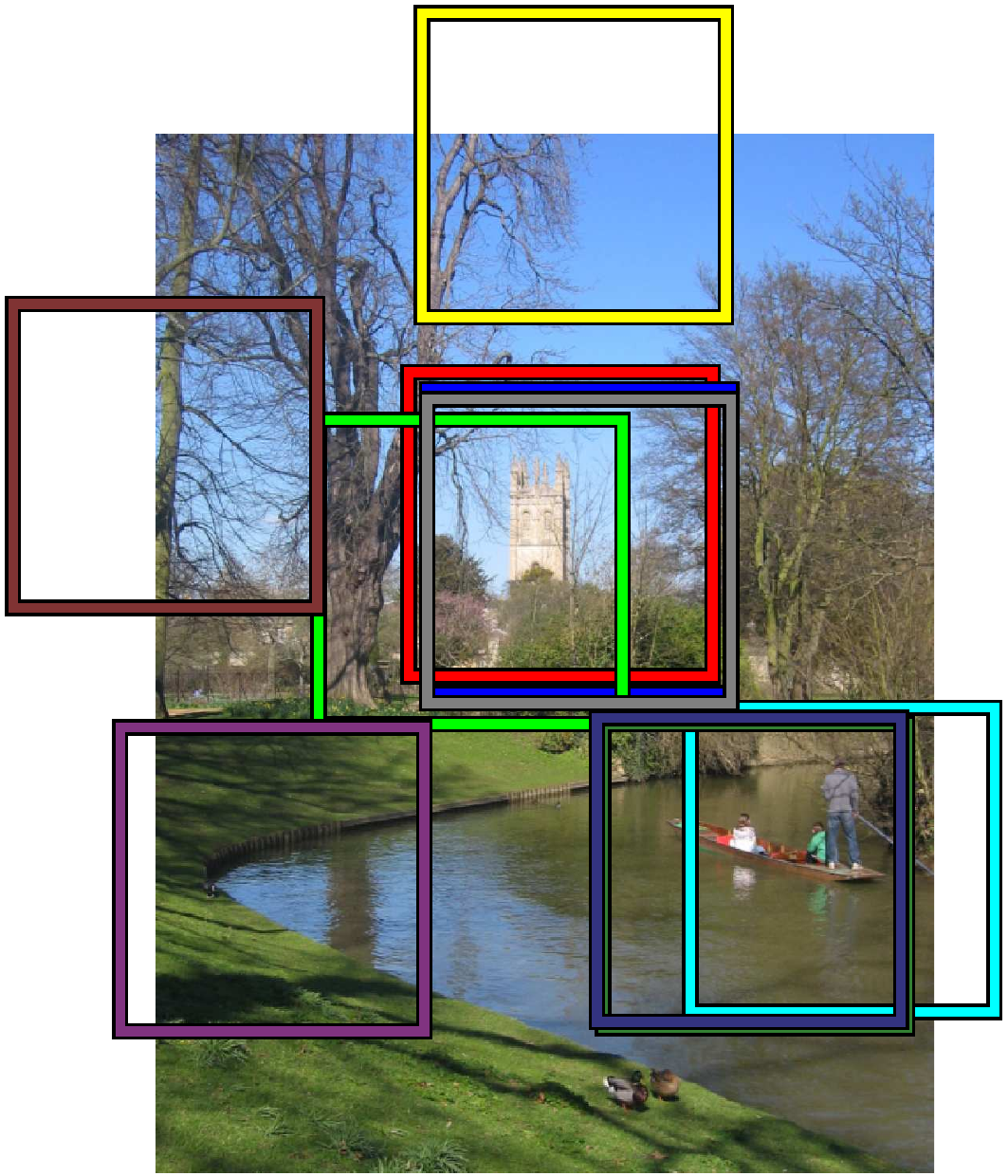}} &
\raisebox{30pt}{
	\begin{tabular}{c}
		\multicolumn{1}{c}{VGG off-the-shelf} \\	
		\foreach \patch in {1,2,3,4,5,6,7,8,9,10}  {
		\includegraphics[height=28pt]{fig/correspondences/q\querytwo_db\dbimagetwo_net0_p\patch_1.png}
		\hspace{-4pt} 
		}\\
		\foreach \patch in {1,2,3,4,5,6,7,8,9,10}  {
		\includegraphics[height=28pt]{fig/correspondences/q\querytwo_db\dbimagetwo_net0_p\patch_2.png} 
		\hspace{-4pt} 
		}\\
	\end{tabular}
}
\\
\includegraphics[height=79pt]{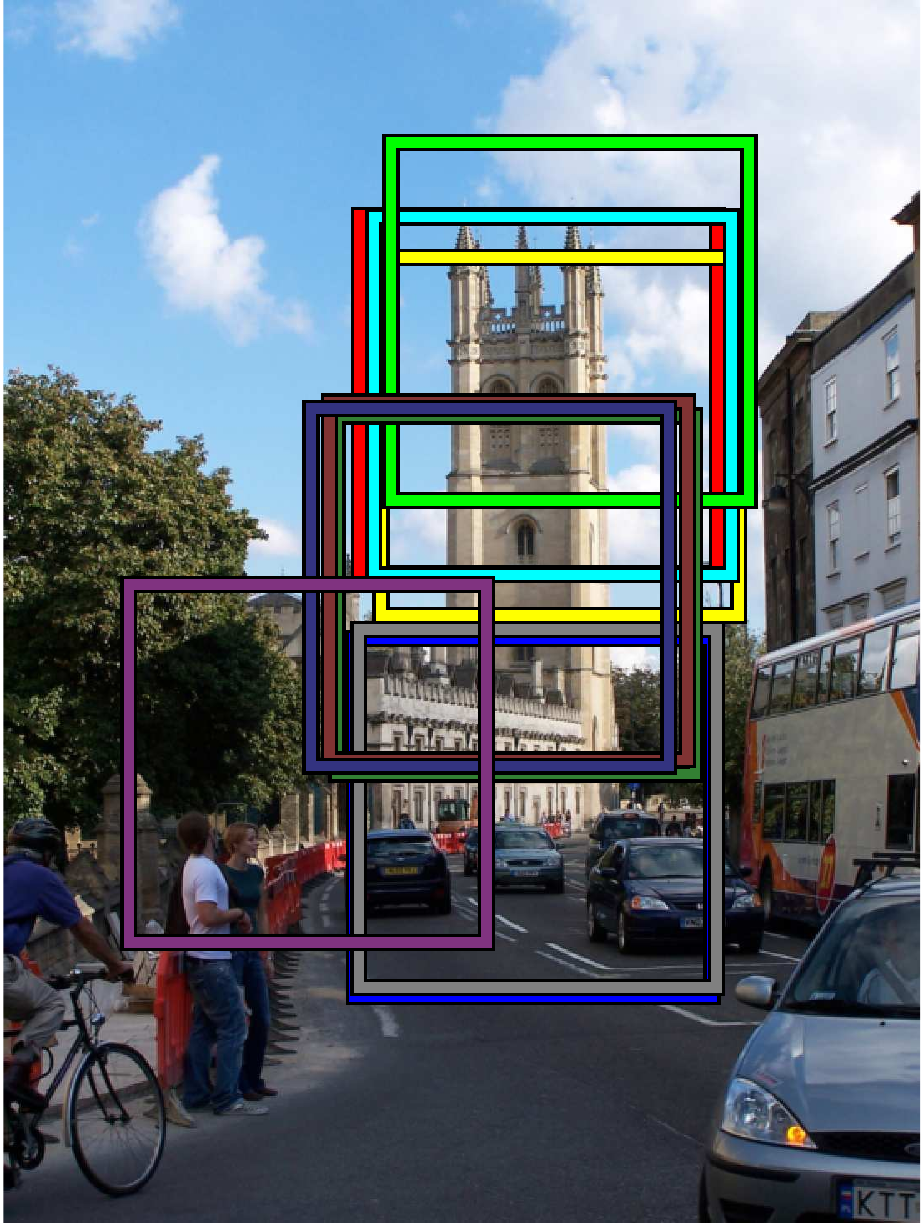} &
~\includegraphics[height=78pt]{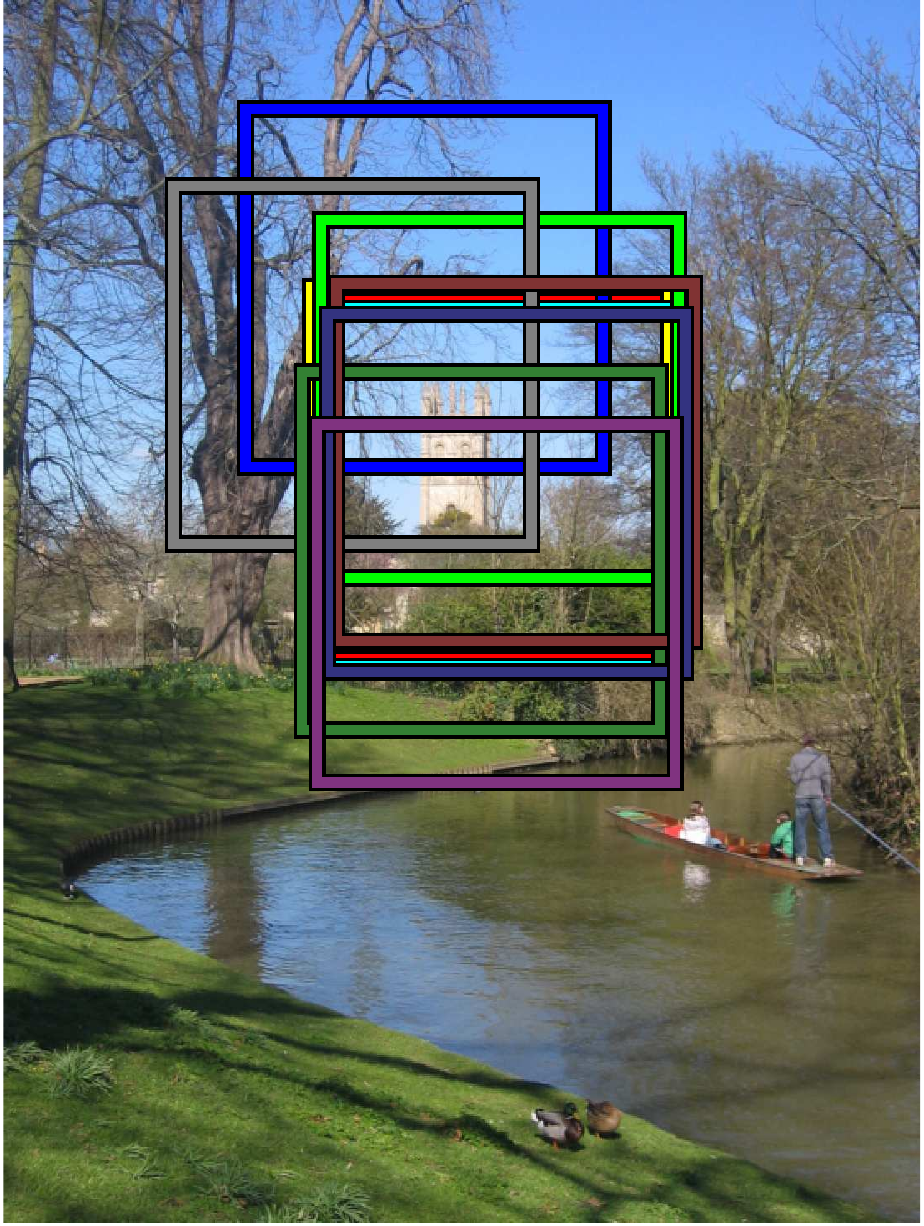} &
\raisebox{30pt}{
	\begin{tabular}{c}
		\multicolumn{1}{c}{VGG ours} \\	
		\foreach \patch in {1,2,3,4,5,6,7,8,9,10}  {
		\includegraphics[height=28pt]{fig/correspondences/q\querytwo_db\dbimagetwo_net1_p\patch_1.png}
		\hspace{-4pt} 
		}\\
		\foreach \patch in {1,2,3,4,5,6,7,8,9,10}  {
		\includegraphics[height=28pt]{fig/correspondences/q\querytwo_db\dbimagetwo_net1_p\patch_2.png} 
		\hspace{-4pt} 
		}\\
	\end{tabular}
}
\\
\end{tabular}
}
%
\caption{Visualization of image regions that correspond to MAC descriptor dimensions that have the highest contribution, \ie large product of descriptor elements, to the pairwise image similarity. The example uses VGG before (top) and after (bottom) fine-tuning. Same color corresponds to the same descriptor component (feature map) per image pair. The patch size is equal to the receptive field of the last local pooling layer.
\label{fig:mac_matches}
\vspace{10pt}
}
\end{figure*}
%
\section{Architecture, learning, search}
\label{sec:network}
In this section we describe the network architecture and present the proposed generalized-pooling layer. 
Then we explain the process of fine-tuning using the contrastive loss and a two-branch network.
We describe how, after fine-tuning, we use the same training data to learn projections that appear to be an effective post-processing step.
Finally, we describe the image representation, search process, and a novel query expansion scheme.
Our proposed architecture is depicted in Figure~\ref{fig:network}.

\subsection{Fully convolutional network}
Our methodology applies to any fully convolutional CNN~\cite{PKS15}.
In practice, popular CNNs for generic object recognition are adopted, such as AlexNet~\cite{KSH12}, VGG~\cite{SZ14}, or ResNet~\cite{HZRS16}, while their fully-connected layers are discarded.
This provides a good initialization to perform the fine-tuning. 

Now, given an input image, the output is a 3D tensor $\cX$ of $W\times H \times K$ dimensions, where $K$ is the number of feature maps in the last layer. 
Let $\cX_k$ be the set of $W\times H$ activations for feature map $k \in \{1 \ldots K$\}.
The network output consists of $K$ such activation sets or 2D feature maps.
We additionally assume that the very last layer is a Rectified Linear Unit (ReLU) such that $\cX$ is non-negative. 

%
\begin{figure*}[t!]
\begin{center}
\def\queryone{32}
\def\dbimageone{5}
\begin{tabular}{c@{\hspace{3pt}}c@{\hspace{20pt}}c@{\hspace{3pt}}c@{\hspace{20pt}}c@{\hspace{3pt}}c}
\includegraphics[height=57pt]{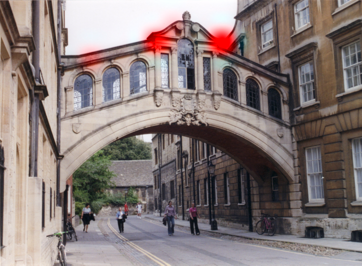}  &
\includegraphics[height=57pt]{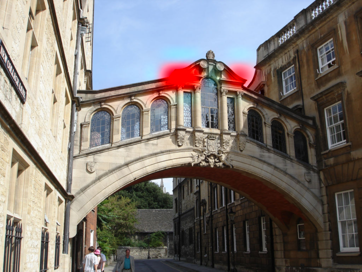}  &
\includegraphics[height=57pt]{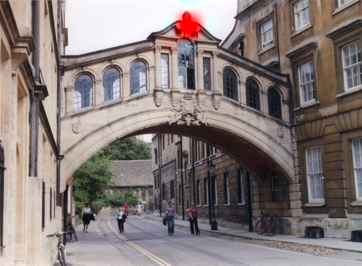}  &
\includegraphics[height=57pt]{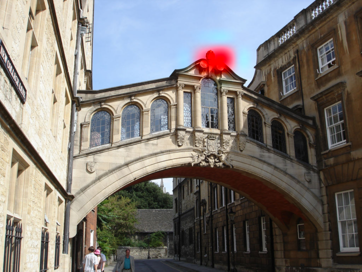}  &
\includegraphics[height=57pt]{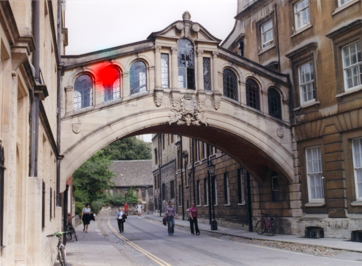}  &
\includegraphics[height=57pt]{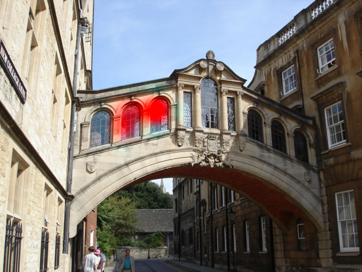}  \\
\multicolumn{2}{c}{$\cX_{469}^p$} & \multicolumn{2}{c}{$\cX_{232}^p$} & \multicolumn{2}{c}{$\cX_{268}^p$} \vspace{2mm} \\
\includegraphics[height=57pt]{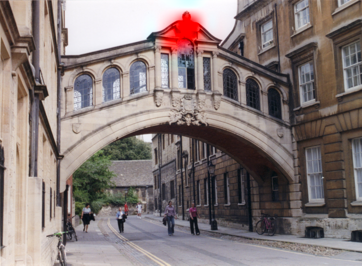}  &
\includegraphics[height=57pt]{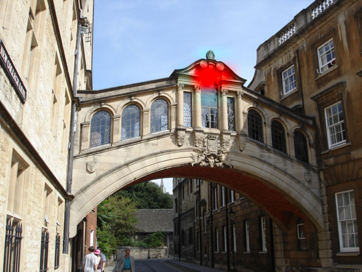}  &
\includegraphics[height=57pt]{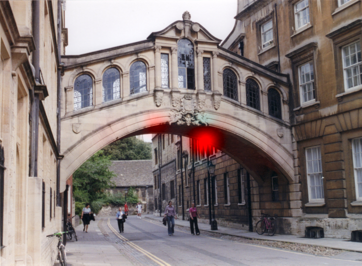}  &
\includegraphics[height=57pt]{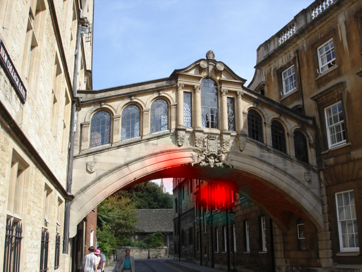}  &
\includegraphics[height=57pt]{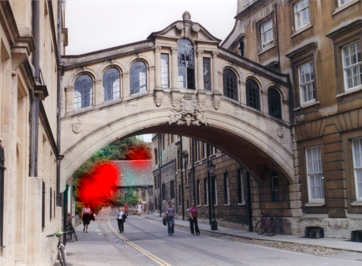}  &
\includegraphics[height=57pt]{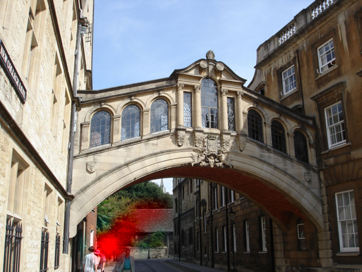}  \\
\multicolumn{2}{c}{$\cX_{99}^p$} & \multicolumn{2}{c}{$\cX_{508}^p$} & \multicolumn{2}{c}{$\cX_{270}^p$} \vspace{2mm} \\
\includegraphics[height=57pt]{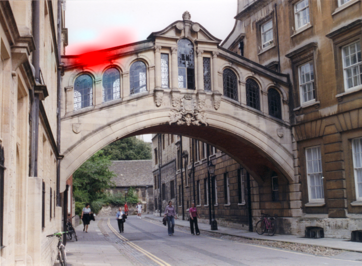}  &
\includegraphics[height=57pt]{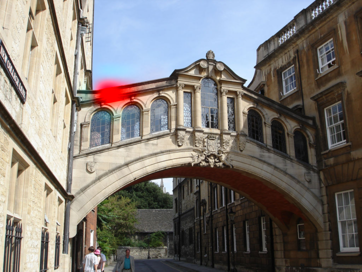}  &
\includegraphics[height=57pt]{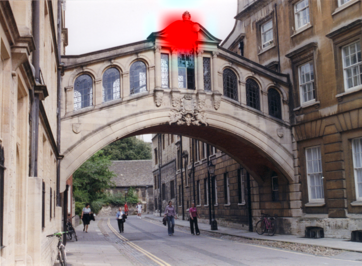}  &
\includegraphics[height=57pt]{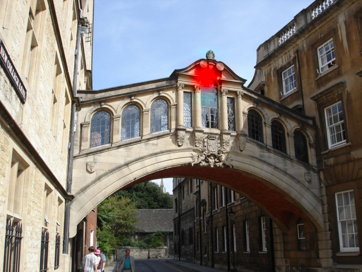}  &
\includegraphics[height=57pt]{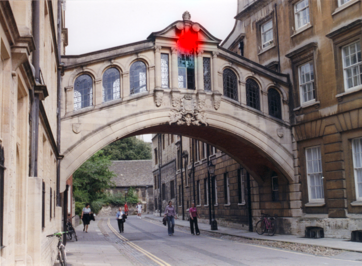}  &
\includegraphics[height=57pt]{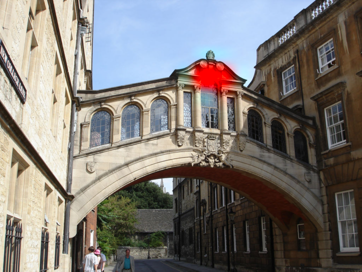}  \\
\multicolumn{2}{c}{$\cX_{436}^p$} & \multicolumn{2}{c}{$\cX_{409}^p$} & \multicolumn{2}{c}{$\cX_{96}^p$} \vspace{2mm} \\
\end{tabular}
\end{center}
\caption{Visualization of $\cX_{k}^p$ projected on the original image for a pair of query-database image. 
The 9 feature maps shown are the ones that score highly, \ie large product of \gem descriptor components, for the database image (right) but low for the top-ranked non-matching images. The example uses fine-tuned VGG with \gem and single $p$ for all feature maps, which converged to 2.92.
\label{fig:gemp_matches}
}
\end{figure*}
%
%
\begin{figure}[b!]
\begin{center}
\def\queryone{42}
\def\dbimageone{211}
\renewcommand{\arraystretch}{1.4}
\begin{tabular}{@{\bsp}c@{\bsp}c@{\bsp}c@{\bsp}}
\includegraphics[height=95pt]{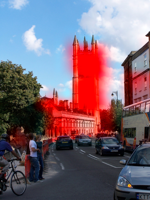}  &
\includegraphics[height=95pt]{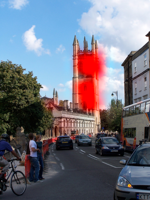}  &
\includegraphics[height=95pt]{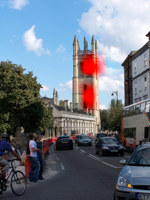} \\
$p=1$ & $p=3$ & $p=10$\\
\end{tabular}
\end{center}
\caption{Visualization of $\cX_{k}^p$ projected on the original image for three different values of $p$. 
Case $p=1$ corresponds to SPoC, and larger $p$ corresponds to \gem before the summation of (\ref{equ:gemp}). 
Examples shown use the off-the-shelf VGG.\label{fig:peffect}}
\end{figure}
%

\subsection{Generalized-mean pooling and image descriptor}
\label{sec:gemp}
We now add a pooling layer that takes $\cX$ as an input and produces a vector $\vf$ as an output of the pooling process.
This vector in the case of the conventional global max pooling (MAC vector~\cite{RSMC14,TSJ16}) is given by 
\begin{equation}
\vf^{(m)} = [\f_1^{(m)} \ldots \f_k^{(m)} \ldots \f_K^{(m)}]^\top \text{,\qquad} \f_k^{(m)} = \max_{x\in \cX_{k}}~x,
\label{equ:maxp}
\end{equation}
while for average pooling (SPoC vector~\cite{BL15}) by
\begin{equation}
\vf^{(a)} = [\f_1^{(a)} \ldots \f_k^{(a)} \ldots \f_K^{(a)}]^\top \text{,\qquad} \f_k^{(a)} = \frac{1}{|\cX_{k}|}\sum_{x\in \cX_{k}}~x.
\label{equ:sump}
\end{equation}
Instead, we exploit the generalized mean~\cite{DTPB09} and propose to use generalized-mean (\gem) pooling  whose result is given by
\begin{equation}
\vf^{(g)} =\! [\f_1^{(g)} \ldots \f_k^{(g)} \ldots \f_K^{(g)}]^\top \text{,\quad} \f_k^{(g)} =\! \left( \frac{1}{|\cX_{k}|}\sum_{x \in \cX_{k}}\!x^{p_k} \right)^\frac{1}{p_k}\text{\hspace{-2ex}.}
\label{equ:gemp}
\end{equation}
Pooling methods~(\ref{equ:maxp})~and~(\ref{equ:sump}) are special cases of \gem pooling given in~(\ref{equ:gemp}), \ie, max pooling when $p_k \rightarrow \infty$ and average pooling for $p_k = 1$.
The feature vector finally consists of a single value per feature map, \ie the generalized-mean activation, and its dimensionality is equal to $K$.
For many popular networks this is equal to 256, 512 or 2048, making it a compact image representation.

The pooling parameter $p_k$ can be manually set or learned since this operation is differentiable and can be part of the back-propagation. 
The corresponding derivatives (while skipping the superscript ${(g)}$ for brevity) are given by

\begin{align}
\frac{\partial\f_k}{\partial x_i} &= \frac{1}{|\cX_{k}|} \f_k^{1-p_k} {x_i}^{p_k-1}, \\[2mm]
\frac{\partial\f_k}{\partial p_k} &= 
\frac{\f_k}{p_k^2}
\left( \log \frac{|\cX_{k}|}{\sum_{x \in \cX_{k}}x^{p_k}} + 
p_k\frac{\sum_{x \in \cX_{k}}x^{p_k}\log x}{\sum_{x \in \cX_{k}}x^{p_k}}\right).
\end{align}
There is a different pooling parameter per feature map in (\ref{equ:gemp}), but it is also possible to use a shared one. In this case $p_k=p,\forall k \in [1, K]$ and we simply denote it by $p$ and not $p_k$. We examine such options in the experimental section and compare to hand-tuned and fixed parameter values.

Max pooling, in the case of MAC, retains one activation per 2D feature map.
In this way, each descriptor component corresponds to an image patch equal to the receptive field.
Then, pairwise image similarity is evaluated via descriptor inner product.
Therefore, MAC similarity implicitly forms patch correspondences.
The strength of each correspondence is given by the product of the associated descriptor components. 
In Figure~\ref{fig:mac_matches} we show the image patches in correspondence that contribute most to the similarity. 
Such implicit correspondences are improved after fine-tuning. Moreover, the CNN fires less on ImageNet classes, \eg cars and bicycles. 

In Figure~\ref{fig:peffect} we show how the spatial distribution of the activations is affected by the generalized mean.
The larger the $p$ the more localized the feature map responses are.
Finally, in Figure~\ref{fig:gemp_matches} we present an example of a query and a database image matched with the fine-tuned VGG with \gem pooling layer (\gem layer in short). We show the feature maps that contribute the most into making this database image being distinguished from non-matching ones that have large similarity, too. 
	
The last network layer comprises an \l2-normalization layer.
Vector $\vf$ is \l2-normalized so that similarity between two images is finally evaluated with inner product. 
In the rest of the paper, \gem vector corresponds to the \l2-normalized vector $\mac$ and constitutes the image descriptor.

\subsection{Siamese learning and loss function}

We adopt a siamese architecture and train a two-branch network. 
Each branch is a clone of the other, meaning that they share the same parameters. 
The training input consists of image pairs $(i,j)$ and labels $Y(i,j)\in \{0, 1\}$ declaring whether a pair is non-matching (label~0) or matching (label~1). 
We employ the contrastive loss~\cite{CHL05} that acts on matching and non-matching pairs and is defined as
\begin{equation}
\loss(i,j) = 
\begin{cases}
\frac{1}{2} ||\mac(i)-\mac(j)||^2, &~\text{if}~Y(i,j) = 1\\
\frac{1}{2}\left(\max\{0, \tau - ||\mac(i)-\mac(j)||\}\right)^2, &~\text{if}~Y(i,j) = 0 \\
\end{cases} 
\end{equation}
where $\mac(i)$ is the \l2-normalized \gem vector of image $i$, and $\tau$ is a margin parameter defining when non-matching pairs have large enough distance in order to be ignored by the loss.
We train the network using a large number of training pairs created automatically (see Section~\ref{sec:dataset}). 
In contrast to other methods~\cite{WSLT+14,SKP15,HA15,AGTPS15}, we find that the contrastive loss generalizes better and converges at higher performance than the triplet loss. 

\subsection{Whitening and dimensionality reduction}
\label{ref:projections}

In this section, the post-processing of fine-tuned \gem vectors is considered.
Previous methods~\cite{BL15,TSJ16} use PCA of an independent set for whitening and dimensionality reduction, \ie the covariance matrix of all descriptors is analyzed. We propose to leverage the labeled data provided by the 3D models and use linear discriminant projections originally proposed by Mikolajczyk and Matas~\cite{MM07}. The projection is decomposed into two parts: whitening and rotation. 
The whitening part is the inverse of the square-root of the intraclass (matching pairs) covariance matrix $C_S^{-\frac{1}{2}}$, where 
\begin{equation}
C_S = \sum_{Y(i,j)=1} \left(\mac(i) - \mac(j)\right)\left(\mac(i) - \mac(j)\right)^\top.
\end{equation}
The rotation part is the PCA of the interclass (non-matching pairs) covariance matrix in the whitened space $\mathrm{eig}(C_S^{-\frac{1}{2}} C_D C_S^{-\frac{1}{2}})$, where 
\begin{equation}
C_D = \sum_{Y(i,j)=0} \left(\mac(i) - \mac(j)\right)\left(\mac(i) - \mac(j)\right)^\top.
\end{equation}
The projection $P = C_S^{-\frac{1}{2}} \mathrm{eig}(C_S^{-\frac{1}{2}} C_D C_S^{-\frac{1}{2}})$ is then applied as $P^\top (\mac(i)-\mu)$, where $\mu$ is the mean \gem vector to perform centering. To reduce the descriptor dimensionality to $D$ dimensions, only eigenvectors corresponding to $D$ largest eigenvalues are used.
Projected vectors are subsequently \l2-normalized.

Our approach uses all available training pairs efficiently in the optimization of the whitening.
It is not optimized in an end-to-end manner and it is performed without using batches of training data.
We first optimize the \gem descriptor and then optimize the whitening.

The described approach acts as a post-processing step, equivalently, once the fine-tuning of the CNN is finished. 
We additionally compare with the end-to-end learning of whitening.
Whitening consists of vector shifting and projection which is modeled in a straightforward manner by a fully connected layer\footnote{The bias is equal to the projected shifting vector.}.
The results favor our approach and are discussed in the experimental section.

\subsection{Image representation and search}
Once the training is finished, an image is fed to the network shown in Figure~\ref{fig:network}. The extracted \gem descriptor is whitened and re-normalized. This constitutes the global descriptor for an image at a single scale. Scale invariance is learned to some extent by the training samples; however, additional invariance is added by multi-scale processing during test time without any additional learning. We follow a standard approach~\cite{GARL16a} and feed the image to the network at multiple scales. The resulting descriptors are finally pooled and re-normalized. This vector constitutes a multi-scale global image representation. We adopt \gem pooling for this state too, which is shown, in our experiments, consistently superior to the standard average pooling. 

Image retrieval is simply performed by exhaustive Euclidean search over database descriptors \wrt the query descriptor. This is equivalent to the inner product evaluation of \l2 normalized vectors, \ie vector-to-matrix multiplication, and sorting. CNN-based descriptors are shown to be highly compatible with approximate-nearest neighbor search methods, in fact, they are very compressible~\cite{GARL16a}. In order to directly evaluate the effectiveness of the learned representation, we do not consider this alternative in this work. In practice, each descriptor requires 4 bytes per dimension to be stored. 

It has recently become a standard policy to combine CNN global image descriptors with simple average query expansion (AQE)~\cite{TSJ16,KMO15,BL15,GARL16a}. An initial query is issued by Euclidean search and AQE acts on the top-ranked nQE images by average pooling of their descriptors. Herein, we argue that tuning nQE to work well across different datasets is not easy. AQE corresponds to a weighted average where nQE descriptors have unit weight and all the rest zero. We generalize this scheme and we propose performing weighted averaging, where the weight of the $i$-th ranked image is given by $(\mac(q)^\top\mac(i))^\alpha$.
The similarity of each retrieved image matters. 
We show in our experiments that AQE is difficult to tune for datasets of different statistics, while this is not the case with the proposed approach.
We refer to this approach as $\alpha$-weighted query expansion ($\alpha$QE).
The proposed $\alpha$QE reduces to AQE for $\alpha=0$.
\section{Training dataset}
\label{sec:dataset}
In this section we briefly summarize the tightly-coupled Bag-of-Words (BoW) image-retrieval and Structure-from-Motion (SfM) 3D reconstruction system~\cite{SRCF15,RSJFCM16} that is employed to automatically select our training data. 
Then, we describe how we use the 3D information to select harder matching pairs and hard non-matching pairs with larger variability. 

\setlength{\fboxsep}{0pt}%
\setlength{\fboxrule}{2pt}%

\begin{figure*}[t]
\centering

\def\imheight{1.6cm}
\def\qnumone{253}
\def\qnumtwo{1611}
\def\qnumthree{1609}

\hspace{-10pt}

\setlength\tabcolsep{4mm}
\renewcommand{\arraystretch}{1.2}

\begin{tabular}{cccc}

\fcolorbox{greenn}{black}{\includegraphics[height=\imheight]{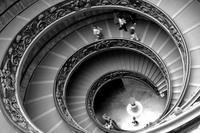}} &
\includegraphics[height=\imheight]{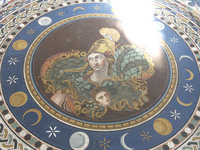} &
\foreach \neg in {2,3}  { 
	\includegraphics[height=\imheight]{fig/negatives/q\qnumone_m2_n\neg.jpg}
} $\ldots$& 
\foreach \neg in {4,5}  { 
	\includegraphics[height=\imheight]{fig/negatives/q\qnumone_m1_n\neg.jpg}
}$\ldots$\\


\fcolorbox{greenn}{black}{\includegraphics[height=\imheight]{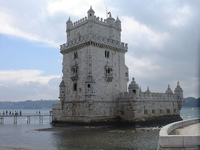}} &
\includegraphics[height=\imheight]{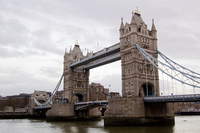} &
\foreach \neg in {3,4}  { 
	\includegraphics[height=\imheight]{fig/negatives/q\qnumtwo_m2_n\neg.jpg}
}$\ldots$ & 
\foreach \neg in {2,5}  { 
	\includegraphics[height=\imheight]{fig/negatives/q\qnumtwo_m1_n\neg.jpg}
}$\ldots$\\ 


\fcolorbox{greenn}{black}{\includegraphics[height=\imheight]{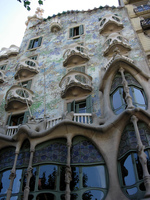}} &
\includegraphics[height=\imheight]{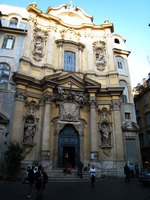} &
\foreach \neg in {2,3,4}  { 
	\includegraphics[height=\imheight]{fig/negatives/q\qnumthree_m2_n\neg.jpg}
}$\ldots$ & 
\foreach \neg in {2,3,4}  { 
	\includegraphics[height=\imheight]{fig/negatives/q\qnumthree_m1_n\neg.jpg}
}$\ldots$\\ \vspace{5pt}

$q$ & $n(q)$ & $\cN_1(q) \setminus  n(q)$ & $\cN_2(q) \setminus  n(q)$ \\ 


\end{tabular}
\caption{Examples of training query image $q$ (one per row shown in green border), and their corresponding negatives chosen by different strategies. We show the hardest non-matching image $n(q)$, and the additional non-matching images selected as negative examples by $\cN_1(q)$ and our  method $\cN_2(q)$. The former chooses k-nearest neighbors among all non-matching images, while the latter chooses k-nearest neighbors but with at most one image per 3D model.
\label{fig:negatives}
}
\end{figure*}

\begin{figure}[b!] 

\def\colsep{\hspace{2mm}}

\def\imheight{1.4}
\def\qnumone{1021}
\def\qnumtwo{1441}
\def\qnumthree{2461}
\def\qnumfour{6511}
\def\qnumfive{3481}
\def\qnumsix{3991}
\def\qnumseven{1111}
\def\qnumeight{5311}

\def\vnum{5}
\def\raisenum{0}

\setlength{\fboxsep}{0pt}%
\setlength{\fboxrule}{2pt}%

\setlength\tabcolsep{0mm}

\renewcommand{\arraystretch}{1.5}
\begin{tabular}{c@{\colsep}c@{\colsep}c@{\colsep}c}
\fcolorbox{greenn}{black}{\includegraphics[height=\imheight cm]{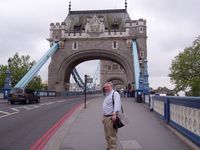}} &
\includegraphics[height=\imheight cm]{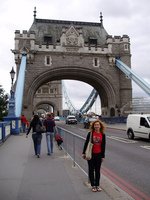}&
\includegraphics[height=\imheight cm]{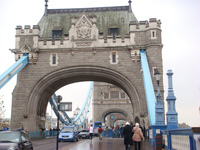}&
\includegraphics[height=\imheight cm]{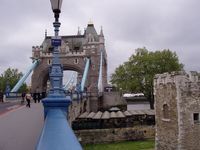}
\\ 
\fcolorbox{greenn}{black}{\includegraphics[height=\imheight cm]{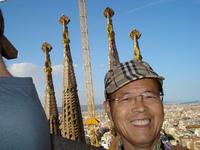}} &
\includegraphics[height=\imheight cm]{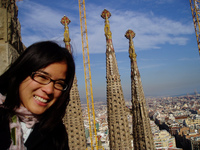}& 
\includegraphics[height=\imheight cm]{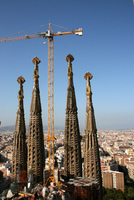}& 
\includegraphics[height=\imheight cm]{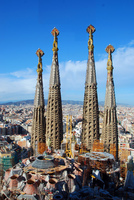}
\\ 
\fcolorbox{greenn}{black}{\includegraphics[height=\imheight cm]{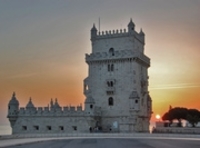}} & 
\includegraphics[height=\imheight cm]{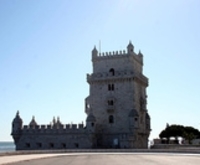}& 
\includegraphics[height=\imheight cm]{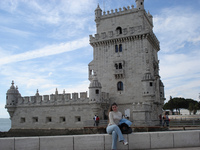}& 
\includegraphics[height=\imheight cm]{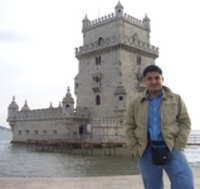}
\\ 
\fcolorbox{greenn}{black}{\includegraphics[height=\imheight cm]{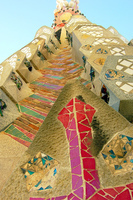}} & 
\includegraphics[height=\imheight cm]{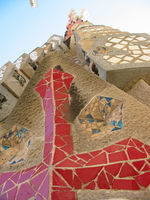} &
\includegraphics[height=\imheight cm]{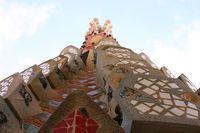}& 
\includegraphics[height=\imheight cm]{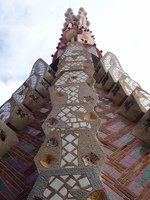}
\\ 
\fcolorbox{greenn}{black}{\includegraphics[height=\imheight cm]{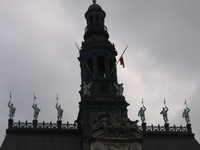}} & 
\includegraphics[height=\imheight cm]{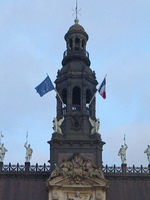} &
\includegraphics[height=\imheight cm]{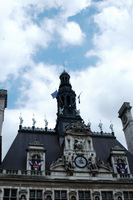} &
\includegraphics[height=\imheight cm]{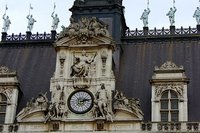}
\\ 
\fcolorbox{greenn}{black}{\includegraphics[height=\imheight cm]{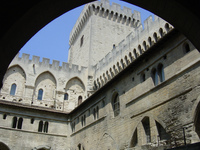}} & 
\includegraphics[height=\imheight cm]{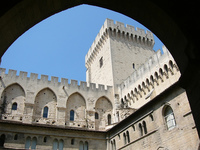} &
\includegraphics[height=\imheight cm]{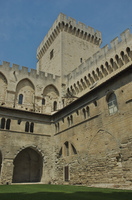} &
\includegraphics[height=\imheight cm]{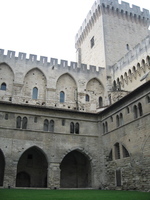}
\\ 
\fcolorbox{greenn}{black}{\includegraphics[height=\imheight cm]{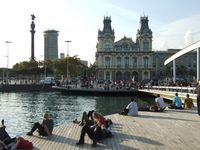}} & 
\includegraphics[height=\imheight cm]{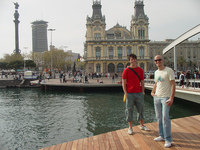} &
\includegraphics[height=\imheight cm]{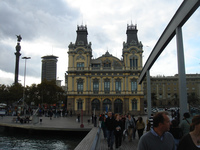} &
\includegraphics[height=\imheight cm]{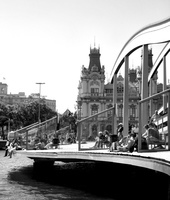}
\\ 
\fcolorbox{greenn}{black}{\includegraphics[height=\imheight cm]{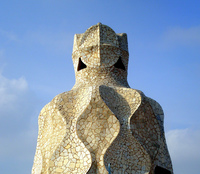}} & 
\includegraphics[height=\imheight cm]{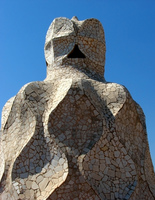} &
\includegraphics[height=\imheight cm]{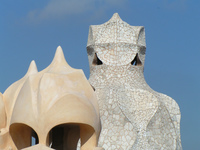} &
\includegraphics[height=\imheight cm]{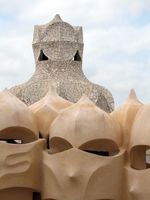}
\\ 
$q$ & $m_1(q)$ & $m_2(q)$ & $m_3(q)$ \\
\end{tabular}
\caption{Examples of training query images (green border) and matching images selected as positive examples by methods: $m_1(q)$~--~the most similar image based on the current network; $m_2(q)$~--~the most similar image based on the BoW representation; and our proposed $m_3(q)$~--~a hard image depicting the same object.
\label{fig:positives}}
\end{figure}

\subsection{BoW and 3D reconstruction}
The retrieval engine used in the work of Schonberger \etal~\cite{SRCF15} builds upon BoW with fast spatial verification~\cite{PCISZ07}. 
It uses Hessian affine local features~\cite{MTSZMSKG05}, \mbox{RootSIFT} descriptors~\cite{AZ12}, and a fine vocabulary of 16M visual words~\cite{MPCM13}.
Then, query images are chosen via min-hash and spatial verification, as in~\cite{CM10a}. 
Image retrieval based on BoW is used to collect images of the objects/landmarks.
These images serve as the initial matching graph for the succeeding SfM reconstruction, which is performed using the state-of-the-art SfM pipeline~\cite{FGGJR10,AFSS+11,SF16}. Different mining techniques, \eg zoom in, zoom out~\cite{MCM13,MRCM14}, sideways crawl~\cite{SRCF15}, help to build a larger and more complete model. 

In this work, we exploit the outcome of such a system. 
Given a large unannotated image collection, images are clustered and a 3D model is constructed per cluster.
We use the terms \emph{3D model}, \emph{model} and \emph{cluster} interchangeably.
For each image, the estimated camera position is known, as well as the local features registered on the 3D model. 
We drop redundant (overlapping) 3D models, that might have been constructed from different seeds.
Models reconstructing the same landmark but from different and disjoint viewpoints are considered as non-overlapping.

\subsection{Selection of training image pairs}
\vspace{3mm}

A 3D model is described as a bipartite visibility graph $\bG = (\cI \cup \cP,\cE)$~\cite{LSH10}, where images $\cI$ and points $\cP$ are the vertices of the graph. 
The edges of this graph are defined by visibility relations between cameras and points, \ie if a point $p\in \cP$ is visible in an image $i\in \cI$, then there exists an edge $(i,p) \in \cE$. 
The set of points observed by an image $i$ is given by
\begin{equation}
\label{equ:observed_points}
\cP(i) = \{ p \in \cP: (i,p) \in \cE \}.
\end{equation} 

We create a dataset of tuples $\left(q, m(q), \cN(q)\right)$, where $q$ represents a query image, $m(q)$ is a positive image that matches the query, and $\cN(q)$ is a set of negative images that do not match the query.
These tuples are used to form training image pairs, where each tuple corresponds to $|\cN(q)|+1$ pairs. 
For a query image $q$, a pool $\cM(q)$ of candidate positive images is constructed based on the camera positions in the cluster of $q$.
It consists of the $k$ images with camera centers closest to the query.
Due to the wide range of camera orientations, these do not necessarily depict the same object. 
We therefore compare three different ways to select the positive image.
The positive examples are fixed during the whole training process for all three strategies.

\vspace{5mm}
\paragraph{Positive images: CNN descriptor distance.} 
The image that has the lowest descriptor distance to the query is chosen as positive, formally
\begin{equation}
m_1(q) = \argmin_{i \in \cM(q)} ||\mac(q)-\mac(i)||.
\label{equ:mac_pos}
\end{equation} 
This strategy is similar to the one followed by Arandjelovic \etal~\cite{AGTPS15}. 
They adopt this choice since only GPS coordinates are available and not camera orientations.
As a consequence, the chosen matching images already have small descriptor distance and, therefore, small loss too.
The network is thus not forced to drastically change/learn by the matching examples, which is the drawback of this approach.

\paragraph{Positive images: maximum inliers.} 
In this approach, the 3D information is exploited to choose the positive image, independently of the CNN descriptor. In particular, the image that has the highest number of co-observed 3D points with the query is chosen.
That is, 
\begin{equation}
\label{equ:ninl_pos}
m_2(q) = \argmax_{i \in \cM(q)} |\cP(q) \cap \cP(i)|.
\end{equation} 
This measure corresponds to the number of spatially verified features between two images, a measure commonly used for ranking in BoW-based retrieval. As this choice is independent of the CNN representation, it delivers more challenging positive examples.

\paragraph{Positive images: relaxed inliers.}
Even though both previous methods choose positive images depicting the same object as the query, the variance of viewpoints is limited.
Instead of using a pool of images with similar camera position, the positive example is selected at random from a set of images that co-observe enough points with the query, but do not exhibit too extreme of a scale change. 
The positive example in this case is  
\begin{equation}
\small
\label{equ:relaxed_pos}
m_3(q) = \texttt{rnd}\left\{ i \in \cM(q): \frac{|\cP(i) \cap \cP(q)|}{|\cP(q)|} \geq t_i,~\texttt{scale}(i,q) \leq t_s \right\},
\end{equation} 
where $\texttt{scale}(i,q)$ is the scale change between the two images.
This method results in selecting harder matching examples that are still guaranteed to depict the same object. Method $m_3$ chooses different image than $m_1$ on 86.5\% of the queries.
In Figure~\ref{fig:positives} we present examples of query images and the corresponding positives selected with the three different methods. The relaxed method increases the variability of viewpoints. 

\paragraph{Negative images.} 
Negative examples are selected from clusters different than the cluster of the query image, as the clusters are non-overlaping. 
We choose hard negatives~\cite{STFKM14,GDDM14}, that is, non-matching images with the most similar descriptor. Two different strategies are proposed: In the first, $\cN_1(q)$, k-nearest neighbors from all non-matching images are selected. In the second, $\cN_2(q)$, the same criterion is used, but at most one image per cluster is allowed. While $\cN_1(q)$ often leads to multiple, and very similar, instances of the same object, $\cN_2(q)$ provides higher variability of the negative examples, see Figure~\ref{fig:negatives}. While positives examples are fixed during the whole training process, hard negatives depend on the current CNN parameters and are re-mined multiple times per epoch. 
\section{Experiments}
\label{sec:experiments}
\vspace{5mm}

In this section we discuss implementation details of our training, evaluate different components of our method, and compare to the state of the art. 

\begin{figure*}[t]
\centering
\input{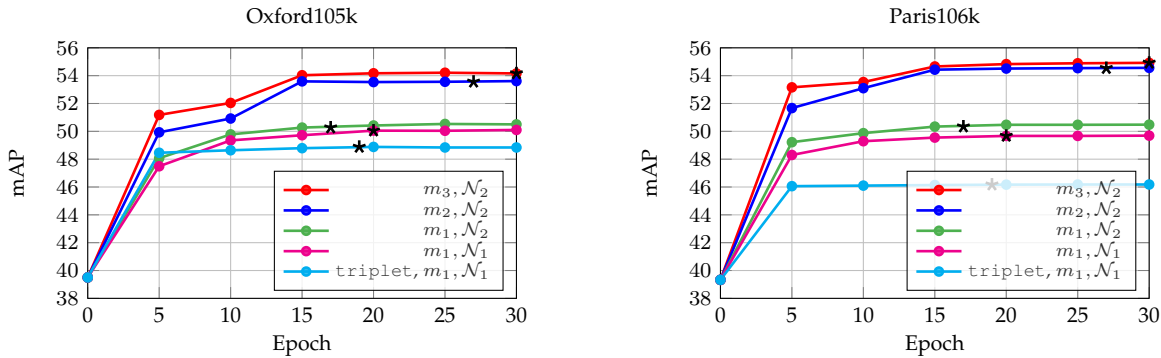}
\begin{tabular}{cc}
\extfig{posnegoxf}{ 
\begin{tikzpicture}
\begin{axis}[%
	font=\footnotesize,
	width=0.4\linewidth,
	height=0.27\linewidth,
	xlabel={Epoch},
	ylabel={mAP},
	title={Oxford105k},
	legend pos=south east,
    legend style={cells={anchor=east}, font =\scriptsize, fill opacity=0.8, row sep=-2.5pt},	
    ymax = 56,
    ymin = 38,
    xmin = 0,
    xmax = 30,
    grid=both,
    xtick={0,5,...,30},    
    xticklabels={0,5,...,30},
    ytick={38, 40, ..., 60},
 	 y label style={at={(axis description cs:.05,.5)}},
 	 x label style={at={(axis description cs:.5,.05)}} 	
]

	\addplot[color=red,     solid, mark=*,  mark size=1.5, line width=1.0] table[x=epoch, y expr={100*\thisrow{siamrelaxed}}]       \posnegoxf;\leg{$m_3,\cN_2$};
	\addplot[color=blue,    solid, mark=*,  mark size=1.5, line width=1.0] table[x=epoch, y expr={100*\thisrow{siamninl}}]          \posnegoxf;\leg{$m_2,\cN_2$};
	\addplot[color=greenn,   solid, mark=*,  mark size=1.5, line width=1.0] table[x=epoch, y expr={100*\thisrow{siammac}}]           \posnegoxf;\leg{$m_1,\cN_2$};
	\addplot[color=magenta, solid, mark=*,  mark size=1.5, line width=1.0] table[x=epoch, y expr={100*\thisrow{siammacnegoff}}]     \posnegoxf;\leg{$m_1,\cN_1$};
	\addplot[color=cyan,    solid, mark=*,  mark size=1.5, line width=1.0] table[x=epoch, y expr={100*\thisrow{tripletmacnegoff}}]  \posnegoxf;\leg{\texttt{triplet}, $m_1,\cN_1$};
	\addplot[color=black, mark=star, only marks, mark size = 2.5, line width = 1] coordinates {(30,54.16)}; 
	\addplot[color=black, mark=star, only marks, mark size = 2.5, line width = 1] coordinates {(27,53.56)};
	\addplot[color=black, mark=star, only marks, mark size = 2.5, line width = 1] coordinates {(17,50.27)};
	\addplot[color=black, mark=star, only marks, mark size = 2.5, line width = 1] coordinates {(20,50.05)};
	\addplot[color=black, mark=star, only marks, mark size = 2.5, line width = 1] coordinates {(19,48.88)};

\end{axis}
\end{tikzpicture}
}
& \hspace{6mm}
\extfig{posnegpar}{ 
\begin{tikzpicture}
\begin{axis}[%
	font=\footnotesize,
	width=0.4\linewidth,
	height=0.27\linewidth,
	xlabel={Epoch},
	ylabel={mAP},
	title={Paris106k},
	legend pos=south east,
    legend style={cells={anchor=east}, font =\scriptsize, fill opacity=0.8, row sep=-2.5pt},	
    ymax = 56,
    ymin = 38,
    xmin = 0,
    xmax = 30,
    grid=both,
    xtick={0,5,...,30},    
    xticklabels={0,5,...,30},
    ytick={38, 40, ..., 60},
 	 y label style={at={(axis description cs:.05,.5)}},
 	 x label style={at={(axis description cs:.5,.05)}} 	    
]
	\addplot[color=red,     solid, mark=*,  mark size=1.5, line width=1.0] table[x=epoch, y expr={100*\thisrow{siamrelaxed}}]       \posnegpar;\leg{$m_3,\cN_2$};
	\addplot[color=blue,    solid, mark=*,  mark size=1.5, line width=1.0] table[x=epoch, y expr={100*\thisrow{siamninl}}]          \posnegpar;\leg{$m_2,\cN_2$};
	\addplot[color=greenn,   solid, mark=*,  mark size=1.5, line width=1.0] table[x=epoch, y expr={100*\thisrow{siammac}}]           \posnegpar;\leg{$m_1,\cN_2$};
	\addplot[color=magenta, solid, mark=*,  mark size=1.5, line width=1.0] table[x=epoch, y expr={100*\thisrow{siammacnegoff}}]     \posnegpar;\leg{$m_1,\cN_1$};
	\addplot[color=cyan,    solid, mark=*,  mark size=1.5, line width=1.0] table[x=epoch, y expr={100*\thisrow{tripletmacnegoff}}]  \posnegpar;\leg{\texttt{triplet}, $m_1,\cN_1$};
	\addplot[color=black, mark=star, only marks, mark size = 2.5, line width = 1] coordinates {(30,54.92)}; 
	\addplot[color=black, mark=star, only marks, mark size = 2.5, line width = 1] coordinates {(27,54.54)};
	\addplot[color=black, mark=star, only marks, mark size = 2.5, line width = 1] coordinates {(17,50.34)};
	\addplot[color=black, mark=star, only marks, mark size = 2.5, line width = 1] coordinates {(20,49.67)};
	\addplot[color=black, mark=star, only marks, mark size = 2.5, line width = 1] coordinates {(19,46.17)};
	
\end{axis}
\end{tikzpicture}
}
\end{tabular}
\vspace{-10pt}
\caption{Performance comparison of methods for positive and negative example selection. Evaluation is performed with AlexNet MAC on Oxford105k and Paris106k datasets. The plot shows the evolution of mAP with the number of training epochs. Epoch 0 corresponds to the off-the-shelf network. All approaches use the contrastive loss, except if otherwise stated. The network with the best performance on the validation set is marked with $\star$.\label{fig:posnegmethod}
\vspace{-7pt}}
\end{figure*}
\begin{figure*}[t]
\centering
\input{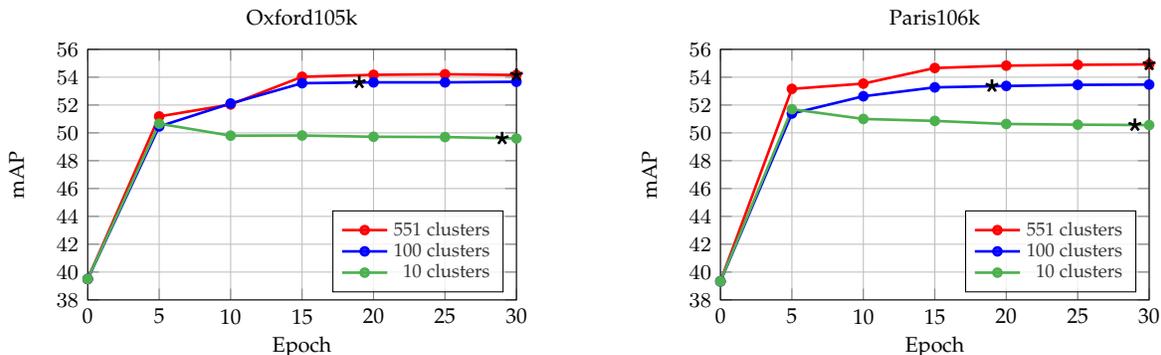}
\begin{tabular}{cc}
\extfig{clustnumoxf}{ 
\begin{tikzpicture}
\begin{axis}[%
	font=\footnotesize,
	width=0.4\linewidth,
	height=0.27\linewidth,
	xlabel={Epoch},
	ylabel={mAP},
	title={Oxford105k},
	legend pos=south east,
    legend style={cells={anchor=east}, font =\scriptsize, fill opacity=0.8, row sep=-2.5pt},	
    ymax = 56,
    ymin = 38,
    xmin = 0,
    xmax = 30,
    grid=both,
    xtick={0,5,...,30},    
    xticklabels={0,5,...,30},
    ytick={38, 40, ..., 60},
 	 y label style={at={(axis description cs:.05,.5)}},
 	 x label style={at={(axis description cs:.5,.05)}} 	    
]
	\addplot[color=red,     solid, mark=*,  mark size=1.5, line width=1.0] table[x=epoch, y expr={100*\thisrow{clustall}}]  \clustnumoxf;\leg{551 clusters};
	\addplot[color=blue,    solid, mark=*,  mark size=1.5, line width=1.0] table[x=epoch, y expr={100*\thisrow{clust100}}]  \clustnumoxf;\leg{100 clusters};
	\addplot[color=greenn,   solid, mark=*,  mark size=1.5, line width=1.0] table[x=epoch, y expr={100*\thisrow{clust10}}]   \clustnumoxf;\leg{10 clusters};
	\addplot[color=black, only marks, mark=star, mark size = 2.5, line width = 1] coordinates {(30,54.16)}; 
	\addplot[color=black, only marks, mark=star, mark size = 2.5, line width = 1] coordinates {(19,53.63)};
	\addplot[color=black, only marks, mark=star, mark size = 2.5, line width = 1] coordinates {(29,49.60)};
	
\end{axis}
\end{tikzpicture}
}
& \hspace{6mm}
\extfig{clustnumpar}{ 
\begin{tikzpicture}
\begin{axis}[%
	font=\footnotesize,
	width=0.4\linewidth,
	height=0.27\linewidth,
	xlabel={Epoch},
	ylabel={mAP},
	title={Paris106k},
	legend pos=south east,
    legend style={cells={anchor=east}, font =\scriptsize, fill opacity=0.8, row sep=-2.5pt},	
    ymax = 56,
    ymin = 38,
    xmin = 0,
    xmax = 30,
    grid=both,
    xtick={0,5,...,30},    
    xticklabels={0,5,...,30},
    ytick={38, 40, ..., 60},
 	 y label style={at={(axis description cs:.05,.5)}},
 	 x label style={at={(axis description cs:.5,.05)}} 	    
]
	\addplot[color=red,     solid, mark=*,  mark size=1.5, line width=1.0] table[x=epoch, y expr={100*\thisrow{clustall}}]  \clustnumpar;\leg{551 clusters};
	\addplot[color=blue,    solid, mark=*,  mark size=1.5, line width=1.0] table[x=epoch, y expr={100*\thisrow{clust100}}]  \clustnumpar;\leg{100 clusters};
	\addplot[color=greenn,   solid, mark=*,  mark size=1.5, line width=1.0] table[x=epoch, y expr={100*\thisrow{clust10}}]   \clustnumpar;\leg{10 clusters};
	\addplot[color=black, mark=star, only marks, mark size = 2.5, line width = 1] coordinates {(30,54.92)}; 
	\addplot[color=black, mark=star, only marks, mark size = 2.5, line width = 1] coordinates {(19,53.37)};
	\addplot[color=black, mark=star, only marks, mark size = 2.5, line width = 1] coordinates {(29,50.56)};
	
\end{axis}
\end{tikzpicture}
}
\end{tabular}
\vspace{-12pt}
\caption{Influence of the number of 3D models used for CNN fine-tuning. Performance is evaluated with AlexNet MAC on Oxford105k and Paris106k datasets using 10, 100 and 551 (all available) 3D models. The network with the best performance on the validation set is marked with $\star$.
\label{fig:clusternumber}
\vspace{-10pt}
}
\end{figure*}

\subsection{Training setup and implementation details}
\paragraph{Structure-from-Motion (SfM).} Our training samples are derived from the dataset used in the work of Schonberger~\etal~\cite{SRCF15}, which consists of 7.4 million images downloaded from Flickr using keywords of popular landmarks, cities and countries across the world.
The clustering procedure~\cite{CM10a} gives around $20$k images to serve as query seeds. 
The extensive retrieval-SfM reconstruction~\cite{RSJFCM16} of the whole dataset results in $1,474$ reconstructed 3D models. 
Removing overlapping models leaves us with $713$ 3D models containing more than $163$k unique images from the initial dataset.
The initial dataset contains, on purpose, all images of Oxford5k and Paris6k datasets. 
In this way, we are able to exclude 98 clusters that contain any image (or their near duplicates) from these test datasets.

\vspace{2mm}
\paragraph{Training pairs.} The size of the 3D models varies from $25$ to $11$k images.
We randomly select $551$ models (around $133$k images) for training and $162$ (around $30k$ images) for validation. 
The number of training queries per 3D model is 10\% of its size and limited to be less or equal to 30.
Around $6,000$ and $1,700$ images are selected for training and validation queries per epoch, respectively.

Each training and validation tuple contains $1$ query, $1$ positive and $5$ negative images.
The pool of candidate positives consists of $k=100$ images with the closest camera centers to the query.
In particular, for method $m_3$, the inlier-overlap threshold is $t_i=0.2$, and the scale-change threshold $t_s=1.5$.
Hard negatives are re-mined $3$ times per epoch, \ie roughly every $2,000$ training queries. 
Given the chosen queries and the chosen positives, we further add 20 images per model to serve as candidate negatives during re-mining.
This constitutes a training set of around $22$k images per epoch when all the training 3D models are used.
The query-tuple selection process is repeated every epoch. This slightly improves the results.

\vspace{2mm}
\paragraph{Learning configuration.} We use MatConvNet~\cite{VL14} for the fine-tuning of networks. 
To perform the fine-tuning as described in Section~\ref{sec:network}, we initialize by the convolutional layers of AlexNet~\cite{KSH12}, VGG16~\cite{SZ14}, or ResNet101~\cite{HZRS16}.
AlexNet is trained using stochastic gradient descent (SGD), while training of VGG and ResNet is more stable with Adam~\cite{KB15}.
We use initial learning rate equal to $l_0 = 10^{-3}$ for SGD, initial stepsize equal to $l_0 = 10^{-6}$ for Adam, an exponential decay $l_0 \exp(-0.1i)$ over epoch $i$, momentum $0.9$, weight decay $5\times 10^{-4}$, margin $\tau$ for contrastive loss $0.7$ for AlexNet, $0.75$ for VGG, and $0.85$ for ResNet, justified by the increase in the dimensionality of the embedding, and a batch size of $5$ training tuples.
All training images are resized to a maximum size of $362 \times 362$, while keeping the original aspect ratio.
Training is done for at most $30$ epochs and the best network is selected based on performance, measured via mean Average Precision (mAP)~\cite{PCISZ07}, on validation tuples. 
Fine-tuning of VGG for one epoch takes around 2 hours on a single TITAN X (Maxwell) GPU with 12 GB of memory.

We overcome GPU memory limitations by associating each query to a tuple, \ie, query plus 6 images (5 positive and 1 negative). Moreover, the whole tuple is processed in the same batch.
Therefore, we feed 7 images to the network, which represents 6 pairs.
In a naive approach, when the query image is different for each pair, 6 pairs require 12 images.

\subsection{Test datasets and evaluation protocol}
\label{sec:exp_test}

\paragraph{Test datasets.} We evaluate our approach on Oxford buildings~\cite{PCISZ07}, Paris~\cite{PCISZ08} and Holidays\footnote{We use the up-right version of Holidays dataset where images are manually rotated so that depicted objects are up-right. This makes us directly comparable to~\cite{GARL16a}. A different version of up-right Holidays is used in our earlier work~\cite{RTC16}, where EXIF metadata is used to rotate the images.}~\cite{JDS08} datasets.
The first two are closer to our training data, while the last is differentiated by containing similar scenes and not only man-made objects or buildings. 
These are also combined with 100k distractors from Oxford100k to allow for evaluation at larger scale. 
The performance is measured via mAP. 
We follow the standard evaluation protocol for Oxford and Paris and crop the query images with the provided bounding box. 
The cropped image is fed as input to the CNN.

\paragraph{Single-scale evaluation.} The dimensionality of the images fed into the CNN is limited to $1024 \times 1024$ pixels.
In our experiments, no vector post-processing is applied if not otherwise stated.

\paragraph{Multi-scale evaluation.} 
Multi-scale representation is only used during test time. 
We resize the input image to different sizes, then feed multiple input images to the network, and finally combine the global descriptors from multiple scales into a single descriptor. 
We compare the baseline average pooling~\cite{GARL16a} with our generalized mean whose pooling parameter is equal to the value learned in the global pooling layer of the network.
In this case, the whitening is learned on the final multi-scale image descriptors.
In our experiments, a single-scale evaluation is used if not otherwise stated. 

\subsection{Results on image retrieval}
\label{sec:exp_results}

\paragraph{Learning.}
We evaluate the off-the-shelf CNN and our fine-tuned ones after different number of training epochs. 
The different methods for positive and negative selection are evaluated independently in order to isolate the benefit of each one. 
Finally, we also perform a comparison with the triplet loss~\cite{AGTPS15}, trained on the same training data as the contrastive loss. Note that a triplet forms two pairs.
Results are presented in Figure~\ref{fig:posnegmethod}.
The results show that positive examples with larger viewpoint variability and negative examples with higher content variability acquire a consistent increase in the performance. 
The triplet loss\footnote{The margin parameter for the triplet loss is set equal to 0.1~\cite{AGTPS15}.} appears to be inferior in our context; we observe oscillation of the error in the validation set from early epochs, which implies over-fitting. 
In the rest of the paper, we adopt the $m_3,\cN_2$ approach.
	
\paragraph{Dataset variability.}
We perform fine-tuning by using a subset of the available 3D models. 
Results are presented in Figure~\ref{fig:clusternumber} with 10, 100 and 551 (all available) clusters, while keeping the amount of training data, \ie number of training queries, fixed.
In the case of 10 and 100 models, we use the largest ones.
It is better to train with all 3D models due to the resulting higher variability in the training set. 
Remarkably, significant increase in performance is achieved even with 10 or 100 models. 
However, the network is able to over-fit in the case of few clusters.
In the rest of our experiments we  use all 551 3D models for training.

\begin{table*}[t!]
\newcolumntype{L}[1]{>{\raggedright\let\newline\\\arraybackslash\hspace{0pt}}m{#1}}
\newcolumntype{C}[1]{>{\centering\let\newline\\\arraybackslash\hspace{0pt}}m{#1}}
\newcolumntype{R}[1]{>{\raggedleft\let\newline\\\arraybackslash\hspace{0pt}}m{#1}}
\newcommand\cw{1.8cm}
\def\arraystretch{1.3}%
\caption{Performance (mAP) comparison after CNN fine-tuning for different pooling layers. 
\gem is evaluated with a single shared pooling parameter or multiple pooling parameters (one for each feature map), which are either fixed or learned. 
A single value or a range is reported in the case of a single or multiple parameters, respectively. 
Results reported with AlexNet and with the use of \cpl2. 
The best performance highlighted in \protect\b{bold}.
\label{tab:pooling}\vspace{-10pt}}
\begin{center}
\setlength{\tabcolsep}{0.0mm}
\footnotesize
\setlength\extrarowheight{1pt}
\begin{tabular}{|@{~~}C{1.5cm}|C{1.5cm}|C{1.5cm}|C{\cw}|C{\cw}|C{\cw}|C{\cw}|C{\cw}|C{\cw}|}
    \hline
    Pooling & Initial p & Learned p &
    Oxford5k & Oxford105k & Paris6k & Paris106k & Holidays & Hol101k \\ \hline\hline
    MAC  & inf & -- & 62.2 & 52.8 & 68.9 & 54.7 & 78.4 & 66.0 \\
    SPoC & 1 & -- & 61.2 & 54.9 & 70.8 & 58.0 & 79.9 & 70.6 \\
    \hline\hline
    \multirow{7}{*}{\gem} 
    & 3 & -- & \b{67.9} & 60.2 & 74.8 & 61.7 & 83.2 & 73.3 \\
    & [2, 5] & -- & 66.8 & 59.7 & 74.1 & 60.8 & \b{84.0} & 73.6 \\
    & [2, 10] & -- & 65.6 & 57.8 & 72.2 & 58.9 & 81.9 & 71.9 \\
    & 3 & 2.32 & 67.7 & \b{60.6} & \b{75.5} & \b{62.6} & 83.7 & \b{73.7} \\ 
    & 3 & [1.0, 6.5] & 66.3 & 57.8 & 74.0 & 60.5 & 83.2 & 72.7 \\
    & [2, 10] & [1.6, 9.9] & 65.3 & 56.4 & 71.4 & 58.6 & 81.4 & 70.8 \\ \hline
\end{tabular}
\end{center}
\end{table*}
\begin{table*}[t!]
\newcolumntype{L}[1]{>{\raggedright\let\newline\\\arraybackslash\hspace{0pt}}m{#1}}
\newcolumntype{C}[1]{>{\centering\let\newline\\\arraybackslash\hspace{0pt}}m{#1}}
\newcolumntype{R}[1]{>{\raggedleft\let\newline\\\arraybackslash\hspace{0pt}}m{#1}}
\newcommand\cw{1.1cm}
\newcommand{\dfs}{\scriptsize} 
\def\arraystretch{1.2}
\caption{Performance (mAP) comparison of CNN vector post-processing: no post-processing, PCA-whitening~\cite{JC12} (\pcawhiten) and our learned whitening (\cpl2). No dimensionality reduction is performed. Fine-tuned AlexNet (Alex) produces a 256D vector and fine-tuned VGG a 512D vector. The best performance highlighted in \protect\b{bold}, the worst in \ww{blue}. The proposed method consistently performs either the best (22 out of 24 cases) or on par with the best method. On the contrary, \pcawhiten~\cite{JC12} often hurts the performance significantly. Best viewed in color.
\label{tab:postproc}}
\begin{center}
\setlength{\tabcolsep}{0.0mm}
\footnotesize
\setlength\extrarowheight{0pt}
\begin{tabular}{|C{1cm}|C{1cm}|C{1cm}|C{\cw}|C{\cw}|C{\cw}|C{\cw}|C{\cw}|C{\cw}|C{\cw}|C{\cw}|C{\cw}|C{\cw}|C{\cw}|C{\cw}|}
    \hline
    \multirow{2}{*}{Net} & \multirow{2}{*}{Post} & \multirow{2}{*}{Dim} &
    \multicolumn{2}{c|}{Oxford5k} & \multicolumn{2}{c|}{Oxford105k} &
    \multicolumn{2}{c|}{Paris6k}  & \multicolumn{2}{c|}{Paris106k} & 
    \multicolumn{2}{c|}{Holidays} & \multicolumn{2}{c|}{Hol101k} \\ 
    \cline{4-15}
    & & & \dfs{MAC} & \dfs{\gem} & \dfs{MAC} & \dfs{\gem} & \dfs{MAC} & \dfs{\gem} & 
    \dfs{MAC} & \dfs{\gem} & \dfs{MAC} & \dfs{\gem} & \dfs{MAC} & \dfs{\gem} \\
    \hline\hline
    \multirow{3}{*}{Alex} & -- & \multirow{3}{*}{256} & 
    60.2 & \ww{60.1} & \b{54.2} & 54.1 & 67.5 & \ww{68.6} & \b{54.9} & \ww{56.9} & 
    \ww{74.5} & \ww{78.7} & 64.8 & \ww{70.9}	\\ 
	& \pcawhiten & &
    \ww{56.9} & 63.7 & \ww{44.1} & \ww{53.7} & \ww{64.3} & 73.2 & \ww{46.8} & 57.4 & 
    75.4 & 82.5 & \ww{63.1} & 71.8 \\ 
	& \cpl2 & &
    \b{62.2} & \b{67.7} & 52.8 & \b{60.6} & \b{68.9} & \b{75.5} & 54.7 & \b{62.6} & 
    \b{78.4} & \b{83.7} & \b{66.0} & \b{73.7} \\ 
	\hline\hline
	\multirow{3}{*}{VGG} & -- & \multirow{3}{*}{512} & 
        82.0 & \ww{82.0} &    76.0 & \ww{76.9} & \ww{78.3} & \ww{79.7} &    71.2 & \ww{72.6} & \ww{79.9} & \ww{83.1} &   \ww{69.4} & \ww{74.5} \\ 
	& \pcawhiten & &
    \ww{78.4} &    83.1 & \ww{71.3} &    77.7 &      80.6 &      84.5 & \ww{70.9} &    76.9 &      82.2 &      86.6 & 70.0 &   75.9 \\ 
	& \cpl2 & &
    \b{82.3} & \b{85.9} & \b{77.0} & \b{81.7} & \b{83.8} & \b{86.0} & \b{76.2} & \b{79.6} & \b{84.1} & \b{87.3} & \b{71.9} & \b{77.1} \\ 
     \hline
\end{tabular}
\end{center}
\end{table*}

\begin{figure*}[t]
\centering
\input{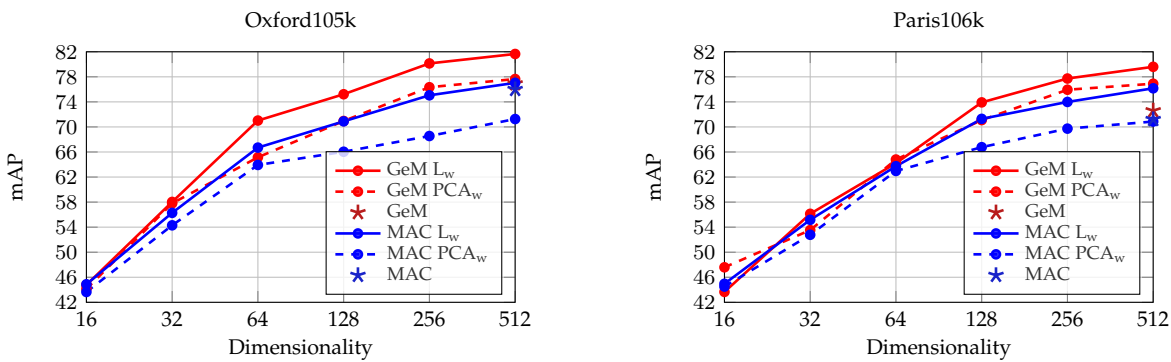}
\begin{tabular}{cc}
\extfig{dimRedOxf}{ 
\begin{tikzpicture}
\begin{semilogxaxis}[%
	font=\footnotesize,
	width=0.4\linewidth,
	height=0.27\linewidth,
	xlabel={Dimensionality},
	ylabel={mAP},
	title={Oxford105k},
	legend pos=south east,
    legend style={cells={anchor=west}, font =\scriptsize, fill opacity=0.8, row sep=-2.5pt},	
    ymax = 82,
    ymin = 42,
    xmin = 16,
    xmax = 512,
    grid=both,
    xtick={8,16,32,64,128,256,512},    
    xticklabels={8,16,32,64,128,256,512},
    ytick={22,26,...,82},
 	 y label style={at={(axis description cs:.05,.5)}},
 	 x label style={at={(axis description cs:.5,.05)}} 	        
]	
	\addplot[color=red,    solid,  mark=*,  mark size=1.5, line width=1.0] table[x=D, y expr={100*\thisrow{gempVGGp}}]  \dimRedOxf;\leg{GeM \cpl2};
	\addplot[color=red,    dashed, mark options={solid}, mark=*,  mark size=1.5, line width=1.0] table[x=D, y expr={100*\thisrow{gempVGGw}}]  \dimRedOxf;\leg{GeM \pcawhiten};
	\addplot[color=darkred, mark=star, only marks, mark size = 3, line width = 1] coordinates {(512,76.9)}; \leg{GeM};
	\addplot[color=blue,    solid,  mark=*,  mark size=1.5, line width=1.0] table[x=D, y expr={100*\thisrow{macVGGp}}]  \dimRedOxf;\leg{MAC \cpl2};
	\addplot[color=blue,    dashed, mark options={solid}, mark=*,  mark size=1.5, line width=1.0] table[x=D, y expr={100*\thisrow{macVGGw}}]  \dimRedOxf;\leg{MAC \pcawhiten};
	\addplot[color=darkblue, mark=star, only marks, mark size = 3, line width = 1] coordinates {(512,76.0)}; \leg{MAC};
	
\end{semilogxaxis}
\end{tikzpicture}
}
& \hspace{6mm}
\extfig{dimRedPar}{ 
\begin{tikzpicture}
\begin{semilogxaxis}[%
	font=\footnotesize,
	width=0.4\linewidth,
	height=0.27\linewidth,
	xlabel={Dimensionality},
	ylabel={mAP},
	title={Paris106k},
	legend pos=south east,
    legend style={cells={anchor=west}, font =\scriptsize, fill opacity=0.8, row sep=-2.5pt},	
    ymax = 82,
    ymin = 42,
    xmin = 16,
    xmax = 512,
    grid=both,
    xtick={8,16,32,64,128,256,512},    
    xticklabels={8,16,32,64,128,256,512},
    ytick={22,26,...,82},
 	 y label style={at={(axis description cs:.05,.5)}},
 	 x label style={at={(axis description cs:.5,.05)}} 	        
]
	\addplot[color=red,    solid,  mark=*,  mark size=1.5, line width=1.0] table[x=D, y expr={100*\thisrow{gempVGGp}}]  \dimRedPar;\leg{GeM \cpl2};
	\addplot[color=red,    dashed, mark options={solid}, mark=*,  mark size=1.5, line width=1.0] table[x=D, y expr={100*\thisrow{gempVGGw}}]  \dimRedPar;\leg{GeM \pcawhiten};
	\addplot[color=darkred, mark=star, only marks, mark size = 3, line width = 1] coordinates {(512,72.6)}; \leg{GeM};
	\addplot[color=blue,    solid,  mark=*,  mark size=1.5, line width=1.0] table[x=D, y expr={100*\thisrow{macVGGp}}]  \dimRedPar;\leg{MAC \cpl2};
	\addplot[color=blue,    dashed, mark options={solid}, mark=*,  mark size=1.5, line width=1.0] table[x=D, y expr={100*\thisrow{macVGGw}}]  \dimRedPar;\leg{MAC \pcawhiten};
	\addplot[color=darkblue, mark=star, only marks, mark size = 3, line width = 1] coordinates {(512,71.2)}; \leg{MAC};
	
\end{semilogxaxis}
\end{tikzpicture}
}
\end{tabular}
\vspace{3pt}
\caption{Performance comparison of the dimensionality reduction performed by \pcawhiten and our \cpl2 with the fine-tuned VGG with MAC layer and the fine-tuned VGG with \gem layer on Oxford105k and Paris106k datasets.
\label{fig:dimreduce}}
\end{figure*}

\paragraph{Pooling methods.}
We evaluate the effect of different pooling layers during CNN fine-tuning.
We present the results in Table~\ref{tab:pooling}.
\gem layer consistently outperforms the conventional max and average pooling. 
This holds for each of the following cases, (i) a single shared pooling parameter $p$ is used, (ii) each feature map has different $p_k$ and (iii) the pooling parameter(s) is (are) either fixed or learned.
Learning a shared parameter turns out to be better than learning multiple ones, as the latter makes the cost function more complex. 
Additionally, the initial values seem to matter to some extent, with a preference for intermediate values. 
Finally, a shared fixed parameter and a shared learned parameter perform similarly, with the latter being slightly better.
This is the case which we adopt for the rest of our experiments, \ie a single shared parameter $p$ that is learned.

\begin{table*}[t!]
\newcolumntype{L}[1]{>{\raggedright\let\newline\\\arraybackslash\hspace{0pt}}m{#1}}
\newcolumntype{C}[1]{>{\centering\let\newline\\\arraybackslash\hspace{0pt}}m{#1}}
\newcolumntype{R}[1]{>{\raggedleft\let\newline\\\arraybackslash\hspace{0pt}}m{#1}}
\newcommand\cw{1.7cm}
\newcommand\sw{0.8cm}
\def\arraystretch{1.3}

\def\om{85.1}  \def\oM{87.9}
\def\obm{80.1} \def\obM{83.3} 
\def\pm{86.0}  \def\pM{89.4}
\def\pbm{79.6} \def\pbM{82.7}
\def\hm{87.3}  \def\hM{91.1}
\def\hbm{77.1} \def\hbM{81.4}

\newcommand{\col}[3]{%
	\newcommand*{\MinNumber}{#2}%
	\newcommand*{\MaxNumber}{#3}%
    \pgfmathsetmacro{\PercentColor}{100.0*(#1-\MinNumber+0.2*(\MaxNumber-\MinNumber))/(\MaxNumber-\MinNumber+0.2*(\MaxNumber-\MinNumber))}%
    \xdef\PercentColor{\PercentColor}%
    \cellcolor{redd!\PercentColor}{#1}
}
\newcommand{\cc}{{\scriptsize{$\blacksquare$}}} 
\newcommand{\sfs}{\footnotesize} 

\caption{Performance (mAP) evaluation of the multi-scale representation using the fine-tuned VGG with \gem layer.
The original scale and down-sampled versions of it are jointly represented.
The pooling parameter used by the generalized mean is the same as the one learned in the \gem layer of the network and equal to 2.92.
The results reported include the use of \cpl2.
\label{tab:multiscale}}
\footnotesize
\setlength\extrarowheight{-1pt}
\begin{center}
\setlength{\tabcolsep}{0mm}
\begin{tabular}{|C{2.8cm}|C{\sw}|C{\sw}|C{\sw}|C{\sw}|C{\sw}|C{\cw}|C{\cw}|C{\cw}|C{\cw}|C{\cw}|C{\cw}|}
    \hline
    \multirow{2}{*}{Pooling over scales} & \multicolumn{5}{c|}{Scale} &
    \multirow{2}{*}{Oxford5k} & \multirow{2}{*}{Oxford105k} & 
    \multirow{2}{*}{Paris6k} & \multirow{2}{*}{Paris106k} & 
    \multirow{2}{*}{Holidays} & \multirow{2}{*}{Hol101k} \\
    \cline{2-6} 
    & \sfs{\sfrac{1}{1}} & \sfs{\sfrac{1}{$\sqrt{2}$}} & \sfs{\sfrac{1}{2}} & 
    \sfs{\sfrac{1}{$\sqrt{8}$}} & \sfs{\sfrac{1}{4}} & & & & & & \\
    \hline\hline
	--    
	& \cc &     &     &     &     & \col{85.9}{\om}{\oM} & \col{81.7}{\obm}{\obM} & \col{86.0}{\pm}{\pM} & \col{79.6}{\pbm}{\pbM} & \col{87.3}{\hm}{\hM} & \col{77.1}{\hbm}{\hbM} \\ \hline\hline
    \multirow{4}{*}{Average}  
    & \cc & \cc &     &     &     & \col{86.8}{\om}{\oM} & \col{82.6}{\obm}{\obM} & \col{86.7}{\pm}{\pM} & \col{80.2}{\pbm}{\pbM} & \col{88.1}{\hm}{\hM} & \col{79.3}{\hbm}{\hbM} \\ 
	& \cc & \cc & \cc &     &     & \col{87.2}{\om}{\oM} & \col{82.4}{\obm}{\obM} & \col{87.3}{\pm}{\pM} & \col{80.6}{\pbm}{\pbM} & \col{89.1}{\hm}{\hM} & \col{79.6}{\hbm}{\hbM} \\ 
	& \cc & \cc & \cc & \cc &     & \col{86.6}{\om}{\oM} & \col{81.9}{\obm}{\obM} & \col{88.2}{\pm}{\pM} & \col{81.3}{\pbm}{\pbM} & \col{89.9}{\hm}{\hM} & \col{79.9}{\hbm}{\hbM} \\ 
    & \cc & \cc & \cc & \cc & \cc & \col{85.1}{\om}{\oM} & \col{80.1}{\obm}{\obM} & \col{88.8}{\pm}{\pM} & \col{81.6}{\pbm}{\pbM} & \col{90.6}{\hm}{\hM} & \col{80.5}{\hbm}{\hbM} \\ \hline\hline
	\multirow{4}{*}{Generalized mean}
    & \cc & \cc &     &     &     & \col{87.3}{\om}{\oM} & \col{83.1}{\obm}{\obM} & \col{86.9}{\pm}{\pM} & \col{80.5}{\pbm}{\pbM} & \col{88.1}{\hm}{\hM} & \col{79.5}{\hbm}{\hbM} \\ 
	& \cc & \cc & \cc &     &     & \col{87.9}{\om}{\oM} & \col{83.3}{\obm}{\obM} & \col{87.7}{\pm}{\pM} & \col{81.3}{\pbm}{\pbM} & \col{89.5}{\hm}{\hM} & \col{79.9}{\hbm}{\hbM} \\ 
	& \cc & \cc & \cc & \cc &     & \col{87.7}{\om}{\oM} & \col{83.2}{\obm}{\obM} & \col{88.7}{\pm}{\pM} & \col{82.3}{\pbm}{\pbM} & \col{89.9}{\hm}{\hM} & \col{80.2}{\hbm}{\hbM} \\ 
	& \cc & \cc & \cc & \cc & \cc & \col{86.8}{\om}{\oM} & \col{82.4}{\obm}{\obM} & \col{89.4}{\pm}{\pM} & \col{82.7}{\pbm}{\pbM} & \col{91.1}{\hm}{\hM} & \col{81.4}{\hbm}{\hbM} \\ \hline
\end{tabular}
\end{center}
\end{table*}

\begin{figure*}[t]
\vspace{-10pt}
\centering
\input{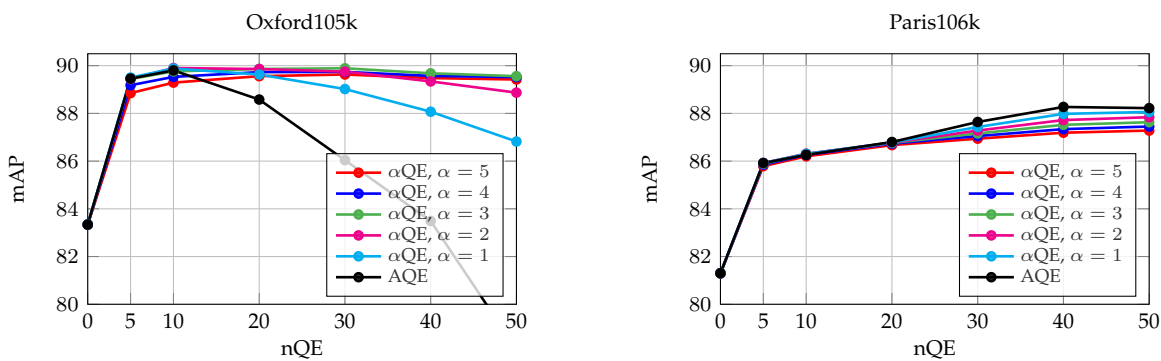}
\begin{tabular}{cc}
\extfig{pQEOxf}{ 
\begin{tikzpicture}
\begin{axis}[%
	font=\footnotesize,
	width=0.4\linewidth,
	height=0.27\linewidth,
	xlabel={nQE},
	ylabel={mAP},
	title={Oxford105k},
	legend pos=south east,
    legend style={cells={anchor=west}, font =\scriptsize, fill opacity=0.8, row sep=-2.5pt},	
    ymax = 90.5,
    ymin = 80,
    xmin = 0,
    xmax = 50,
    grid=both,
    xtick={0,5,10,20,...,50},    
    xticklabels={0,5,10,20,...,50},
    ytick={78, 80, ..., 92},
 	 y label style={at={(axis description cs:.05,.5)}},
 	 x label style={at={(axis description cs:.5,.05)}} 	
]

	\addplot[color=red,     solid, mark=*,  mark size=1.5, line width=1.0] table[x=nQE, y expr={100*\thisrow{p5}}] \pQEOxf;\leg{$\alpha$QE, $\alpha=5$};
	\addplot[color=blue,    solid, mark=*,  mark size=1.5, line width=1.0] table[x=nQE, y expr={100*\thisrow{p4}}] \pQEOxf;\leg{$\alpha$QE, $\alpha=4$};
	\addplot[color=greenn,  solid, mark=*,  mark size=1.5, line width=1.0] table[x=nQE, y expr={100*\thisrow{p3}}] \pQEOxf;\leg{$\alpha$QE, $\alpha=3$};
	\addplot[color=magenta, solid, mark=*,  mark size=1.5, line width=1.0] table[x=nQE, y expr={100*\thisrow{p2}}] \pQEOxf;\leg{$\alpha$QE, $\alpha=2$};
	\addplot[color=cyan,    solid, mark=*,  mark size=1.5, line width=1.0] table[x=nQE, y expr={100*\thisrow{p1}}] \pQEOxf;\leg{$\alpha$QE, $\alpha=1$};
	\addplot[color=black,   solid, mark=*,  mark size=1.5, line width=1.0] table[x=nQE, y expr={100*\thisrow{p0}}] \pQEOxf;\leg{AQE};

\end{axis}
\end{tikzpicture}
}
& \hspace{6mm}
\extfig{pQEPar}{ 
\begin{tikzpicture}
\begin{axis}[%
	font=\footnotesize,
	width=0.4\linewidth,
	height=0.27\linewidth,
	xlabel={nQE},
	ylabel={mAP},
	title={Paris106k},
	legend pos=south east,
    legend style={cells={anchor=west}, font =\scriptsize, fill opacity=0.8, row sep=-2.5pt},	
    ymax = 90.5,
    ymin = 80,
    xmin = 0,
    xmax = 50,
    grid=both,
    xtick={0,5,10,20,...,50},    
    xticklabels={0,5,10,20,...,50},
    ytick={78, 80, ..., 92},
 	 y label style={at={(axis description cs:.05,.5)}},
 	 x label style={at={(axis description cs:.5,.05)}} 	
]

	\addplot[color=red,     solid, mark=*,  mark size=1.5, line width=1.0] table[x=nQE, y expr={100*\thisrow{p5}}] \pQEPar;\leg{$\alpha$QE, $\alpha=5$};
	\addplot[color=blue,    solid, mark=*,  mark size=1.5, line width=1.0] table[x=nQE, y expr={100*\thisrow{p4}}] \pQEPar;\leg{$\alpha$QE, $\alpha=4$};
	\addplot[color=greenn,  solid, mark=*,  mark size=1.5, line width=1.0] table[x=nQE, y expr={100*\thisrow{p3}}] \pQEPar;\leg{$\alpha$QE, $\alpha=3$};
	\addplot[color=magenta, solid, mark=*,  mark size=1.5, line width=1.0] table[x=nQE, y expr={100*\thisrow{p2}}] \pQEPar;\leg{$\alpha$QE, $\alpha=2$};
	\addplot[color=cyan,    solid, mark=*,  mark size=1.5, line width=1.0] table[x=nQE, y expr={100*\thisrow{p1}}] \pQEPar;\leg{$\alpha$QE, $\alpha=1$};
	\addplot[color=black,   solid, mark=*,  mark size=1.5, line width=1.0] table[x=nQE, y expr={100*\thisrow{p0}}] \pQEPar;\leg{AQE};

\end{axis}
\end{tikzpicture}
}
\end{tabular}
\vspace{-2pt}
\caption{Performance evaluation of our $\alpha$-weighted query expansion ($\alpha$QE) with the VGG with \gem layer, multi-scale representation, and \cpl2 on Oxford105k and Paris106k datasets.
We compare the standard average query expansion (AQE) to our $\alpha$QE for different values of $\alpha$ and number of images used nQE. 
\label{fig:pQE}
\vspace{-7pt}}
\end{figure*}
\begin{table}[t!]
\newcolumntype{L}[1]{>{\raggedright\let\newline\\\arraybackslash\hspace{0pt}}m{#1}}
\newcolumntype{C}[1]{>{\centering\let\newline\\\arraybackslash\hspace{0pt}}m{#1}}
\newcolumntype{R}[1]{>{\raggedleft\let\newline\\\arraybackslash\hspace{0pt}}m{#1}}
\newcommand\cw{1.1cm}
\def\arraystretch{1.3}

\caption{Performance (mAP) evaluation for varying descriptor dimensionality after reduction with \cpl2.
Results reported with the fine-tuned VGG with \gem and the fine-tuned ResNet (Res) with \gem. 
Multi-scale representation is used at the test time for both networks.
\label{tab:dimreduction}\vspace{-10pt}}
\footnotesize
\setlength\extrarowheight{1pt}
\begin{center}
\setlength{\tabcolsep}{0mm}
\begin{tabular}{|C{0.8cm}|@{\ssp}r@{\ssp}|C{\cw}|C{\cw}|C{\cw}|C{\cw}|C{\cw}|C{\cw}|}
    \hline
    Net & Dim & Oxf5k & Oxf105k & Par6k & Par106k & Hol & Hol101k \\
    \hline\hline
    \multirow{7}{*}{VGG} 
    & 512 & 87.9 & 83.3 & 87.7 & 81.3 & 89.5 & 79.9 \\
    & 256 & 85.4 & 79.7 & 85.7 & 78.2 & 87.8 & 77.2 \\
    & 128 & 81.6 & 75.4 & 83.4 & 74.9 & 84.4 & 72.6 \\
    & 64  & 77.0 & 69.9 & 77.4 & 66.7 & 81.1 & 66.2 \\
    & 32  & 66.9 & 57.4 & 72.2 & 58.6 & 72.9 & 54.3 \\
    & 16  & 56.2 & 44.4 & 63.5 & 45.5 & 60.9 & 36.9 \\
    & 8   & 34.1 & 25.7 & 43.9 & 29.0 & 43.4 & 13.8 \\
    \hline\hline
    \multirow{9}{*}{Res}
    & 2048 & 87.8 & 84.6 & 92.7 & 86.9 & 93.9 & 87.9 \\
    & 1024 & 86.2 & 82.4 & 91.8 & 85.3 & 92.5 & 86.1 \\
    & 512 & 84.6 & 80.4 & 90.0 & 82.6 & 90.6 & 83.2 \\
    & 256 & 83.1 & 77.3 & 87.5 & 78.8 & 88.4 & 80.2 \\
    & 128 & 79.5 & 72.2 & 84.5 & 74.3 & 85.9 & 76.5 \\
    & 64 & 74.0 & 65.8 & 78.4 & 65.3 & 80.3 & 66.9 \\
    & 32 & 57.9 & 48.5 & 70.8 & 56.1 & 71.2 & 51.9 \\
    & 16 & 40.3 & 31.8 & 61.8 & 45.6 & 56.4 & 31.3 \\
    & 8 & 25.3 & 16.3 & 44.3 & 27.8 & 37.8 & 11.4 \\
    \hline
\end{tabular}
\end{center}
\end{table}

\paragraph{Learned projections.}
The PCA-whitening~\cite{JC12} (\pcawhiten) is shown to be essential in some cases of CNN-based descriptors~\cite{BSCL14,BL15,TSJ16}.
On the other hand, it is shown that on some datasets, the performance after \pcawhiten substantially drops compared to the raw descriptors (max pooling on Oxford5k~\cite{BL15}). 
We perform comparison of the traditional whitening methods and the proposed learned discriminative whitening (\cpl2), described in Section~\ref{ref:projections}.
Table~\ref{tab:postproc} shows results without post-processing, with \pcawhiten and with \cpl2.
Our experiments confirm that \pcawhiten often reduces the performance. 
In contrast to that, the proposed \cpl2 achieves the best performance in most cases and is never the worst-performing method. 
Compared with the no post-processing baseline, \cpl2 reduces the performance twice for AlexNet, but the drop is negligible compared to the drop observed for \pcawhiten. For VGG, the proposed \cpl2 {\em always} outperforms the no post-processing baseline.

We conduct an additional experiment by appending a whitening layer at the end of the network during fine-tuning.
In this way, whitening is learned in an end-to-end manner, along with the convolutional filters and with the same training data in batch-mode. 
Dropout~\cite{SHKSS14} is additionally used for this layer which we find to be essential.
We observe that convergence of the network comes much slower in this case, \ie after 60 epochs. 
Moreover, the final achieved performance is not higher than our \cpl2. In particular, end-to-end whitening on AlexNet MAC achieves 49.6 and 52.1 mAP on Oxford105k and Paris106k, respectively, while our \cpl2 on the same network achieves 52.8 and 54.7 mAP on Oxford105k and Paris106k, respectively. 
Therefore, we adopt \cpl2 as it is much faster to train and more effective.

\vspace{2mm}
\paragraph{Dimensionality reduction.}
We compare dimensionality reduction performed with \pcawhiten~\cite{JC12} and with our \cpl2. The performance for varying descriptor dimensionality is plotted in Figure~\ref{fig:dimreduce}. The plots suggest that \cpl2 works better in most dimensionalities.

\begin{table*}[h!]
\newcommand{\ebA}{\texttt{\bf{A}}}
\newcommand{\ebV}{\texttt{\bf{V}}}
\newcommand{\ebf}{{\scriptsize{$\blacksquare$}}}
\newcommand{\our}{$\boldsymbol{\star}$\xspace}
\newcolumntype{L}[1]{>{\raggedright\let\newline\\\arraybackslash\hspace{0pt}}m{#1}}
\newcolumntype{C}[1]{>{\centering\let\newline\\\arraybackslash\hspace{0pt}}m{#1}}
\newcolumntype{R}[1]{>{\raggedleft\let\newline\\\arraybackslash\hspace{0pt}}m{#1}}
\def\arraystretch{1.3}
\newcommand\cw{1.6cm}
\caption{Performance (mAP) comparison with the state-of-the-art image retrieval using VGG and ResNet (Res) deep networks, and using local features. 
F-tuned: Use of the fine-tuned network (yes), or the off-the-shelf network~(no), not applicable for the methods using local features (n/a). 
Dim:~Dimensionality of the final compact image representation, not applicable (n/a) for the BoW based methods due to their sparse representation. 
Our methods are marked with \our and they are always accompanied by the multi-scale representation and our learned whitening \cpl2.\newline
Previous state of the art is highlighted in \protect\ob{bold}, new state of the art in \protect\nb{red outline}. 
Best viewed in color.
\label{tab:stateofart}}
\footnotesize
\setlength\extrarowheight{-1pt}
\begin{center}
\setlength{\tabcolsep}{0.2mm}
\begin{tabular}{|C{1.1cm}|@{~~}L{3.6cm}|C{1.1cm}|C{1.1cm}|C{\cw}|C{\cw}|C{\cw}|C{\cw}|C{\cw}|C{\cw}|}
    \hline
    Net & Method & F-tuned & Dim & 
    Oxford5k & Oxford105k & Paris6k & Paris106k & Holidays & Hol101k \\ 
    \hline\hline
    \multicolumn{10}{|c|}{\b{Compact representation using deep networks}} \\
    \hline\hline
    \multirow{11}{*}{VGG} 
    & MAC\cite{RSMC14}$^\dagger$ & no & 512 & 56.4 & 47.8 & 72.3 & 58.0 & 79.0 & 66.1 \\
    & SPoC~\cite{BL15}$^\dagger$ & no & 512 & 68.1 & 61.1 & 78.2 & 68.4 & 83.9 & \ob{75.1} \\
    & CroW~\cite{KMO15} & no & 512 & 70.8 & 65.3 & 79.7 & 72.2 & 85.1 & -- \\    
    & R-MAC~\cite{TSJ16} & no & 512 & 66.9 & 61.6 &  83.0 & 75.7 & {~~}86.9$^\ddagger$ & -- \\
    & BoW-CNN~\cite{MMOS+16} & no & n/a & 73.9 & 59.3 & 82.0 & 64.8 & -- & -- \\
    & NetVLAD~\cite{AGTPS15} & no & 4096 & 66.6 & -- & 77.4 & -- & 88.3 & --  \\
    & NetVLAD~\cite{AGTPS15} & yes & 512 & 67.6 & -- & 74.9 & -- & 86.1 & -- \\
    & NetVLAD~\cite{AGTPS15} & yes & 4096 & 71.6 & -- & 79.7 & -- & 87.5 & -- \\
    & Fisher Vector~\cite{OHB16} & yes & 512 & 81.5 & 76.6 & 82.4 & -- & -- & -- \\
    & R-MAC~\cite{GARL16} & yes & 512 & \ob{83.1} & \ob{78.6} & \ob{87.1} & \ob{79.7} & \ob{89.1} & -- \\
    & \our \gem & yes & 512 & \nb{87.9} & \nb{83.3} & \nb{87.7} & \nb{81.3} & \nb{89.5} & \nb{79.9} \\
    \hline\hline
    \multirow{3}{*}{Res} 
    & R-MAC~\cite{TSJ16}$^\ddagger$ & no & 2048 & 69.4 & 63.7 & 85.2 & 77.8 & 91.3 & -- \\
    & R-MAC~\cite{GARL16a} & yes & 2048 & \ob{86.1} & \ob{82.8} & \ob{94.5} & \ob{90.6} & \ob{94.8} & -- \\
    & \our \gem & yes & 2048 & \nb{87.8} & \nb{84.6} & 92.7 & 86.9 & 93.9 & \nb{87.9} \\
    \hline\hline
    \multicolumn{10}{|c|}{\b{Re-ranking (R) and query expansion (QE)}} \\
    \hline\hline
    \multirow{3}{*}{n/a} 
    & BoW+R+QE~\cite{CMPM11} & n/a & n/a & 82.7 & 76.7 & 80.5 & 71.0 & -- & -- \\
    & BoW-fVocab+R+QE~\cite{MPCM13} & n/a & n/a & 84.9 & 79.5 & 82.4 & 77.3 & 75.8 & -- \\
    & HQE~\cite{TJ14} & n/a & n/a & 88.0 & 84.0 & 82.8 & -- & -- & -- \\
    \hline\hline
    \multirow{5}{*}{VGG} 
    & CroW+QE~\cite{KMO15} & no & 512 & 74.9 & 70.6 & 84.8 & 79.4 & -- & -- \\
    & R-MAC+R+QE~\cite{TSJ16} & no & 512 & 77.3 & 73.2 & 86.5 & 79.8 & -- & -- \\ 
    & BoW-CNN+R+QE~\cite{MMOS+16} & no & n/a & 78.8 & 65.1 & 84.8 & 64.1 & -- & -- \\
    & R-MAC+QE~\cite{GARL16} & yes & 512 & \ob{89.1} & \ob{87.3} & \ob{91.2} & \ob{86.8} & -- & -- \\
    & \our \gem+$\alpha$QE & yes & 512 & \nb{91.9} & \nb{89.6} & \nb{91.9} & \nb{87.6} & -- & -- \\
    \hline\hline
    \multirow{3}{*}{Res} 
    & R-MAC+QE~\cite{TSJ16}$^\ddagger$ & no & 2048 & 78.9 & 75.5 & 89.7 & 85.3 & -- & -- \\
    & R-MAC+QE~\cite{GARL16a} & yes & 2048 & \ob{90.6} & \ob{89.4} & \ob{96.0} & \ob{93.2} & -- & -- \\
    & \our \gem+$\alpha$QE & yes & 2048 & \nb{91.0} & \nb{89.5} & 95.5 & 91.9 & -- & -- \\
    \hline
\end{tabular}
\end{center}
\scriptsize
$~~~~~~~^\dagger$:~Our evaluation of MAC and SPoC with \pcawhiten and with the off-the-shelf network. \\ 
$~~~~~~~~^\ddagger$:~Evaluation of R-MAC by~\cite{GARL16a} with the off-the-shelf network.
\end{table*}

\paragraph{Multi-scale representation.}
We evaluate multi-scale representation constructed at test time without any additional learning.
We compare the previously used averaging of descriptors at multiple image scales~\cite{GARL16a} with our generalized-mean of the same descriptors.
Results are presented in Table~\ref{tab:multiscale}, where there is a significant benefit when using the multi-scale \gem.
It also offers some improvement over average pooling.
In the rest of our experiments we adopt multi-scale representation, pooled by generalized mean, for scales 1, $\nicefrac{1}{\sqrt{2}}$, and $\nicefrac{1}{2}$.
Results using the supervised dimensionality reduction by \cpl2 on the multi-scale \gem representation are shown in Table~\ref{tab:dimreduction}.

\vspace{2mm}
\paragraph{Query expansion.}
We evaluate the proposed $\alpha$QE, which reduces to AQE for $\alpha=0$, and present results in Figure~\ref{fig:pQE}.
Note that Oxford and Paris have different statistics in terms of the number of relevant images per query. The average, minimum, and maximum number of positive images per query on Oxford is 52, 6, and 221, respectively. The same measurements for Paris are 163, 51, and 289. 
As a consequence, AQE behaves in a very different way across these dataset, while our $\alpha$QE is a more stable choice. We finally set $\alpha=3$ and $\text{nQE}=50$.

\paragraph{Over-fitting and generalization.}
In all experiments, all clusters including any image (not only query landmarks) from Oxford5k or Paris6k datasets are removed. 
We now repeat the training using all 3D models, including those of Oxford and Paris landmarks.
In this way, we evaluate whether the network tends to over-fit to the training data or to generalize.
The same amount of training queries is used for a fair comparison.
We observe negligible difference in the performance of the network on Oxford and Paris evaluation results, \ie the difference in mAP was on average $+0.3$ over all testing datasets. 
We conclude that the network generalizes well and is relatively insensitive to over-fitting. 

\paragraph{Comparison with the state of the art.}
We extensively compare our results with the state-of-the-art performance on compact image representations and on approaches that do query expansion. 
The results for the fine-tuned \gem based networks are summarized together with previously published results in Table~\ref{tab:stateofart}.
The proposed methods outperform the state of the art on all datasets when the VGG network architecture and initialization are used.
Our method is outperformed by the work of Gordo \etal on Paris with the ResNet architecture, while we have the state-of-the-art score on Oxford. We are on par with the state-of-the-art on Holidays. 
Note, however, that we did not perform any manual labeling or cleaning of our training data, while in their work landmark labels were used.
We additionally combine \gem with query expansion and further boost the performance.

\section{Conclusions}
We addressed fine-tuning of CNN for image retrieval. The training data are selected from an automated 3D reconstruction system applied on a large unordered photo collection. 
The reconstructions consist of buildings and popular landmarks; however, the same process is applicable to any rigid 3D objects.
The proposed method does not require any manual annotation and yet achieves top performance on standard benchmarks. 
The achieved results reach the level of the best systems based on local features with spatial matching and query expansion while being faster and requiring less memory. 
The proposed pooling layer that generalizes previously adopted mechanisms is shown to improve the retrieval accuracy while it is also effective for constructing a joint multi-scale representation.
Training data, trained models, and code are publicly available.

\ifCLASSOPTIONcompsoc
  \section*{Acknowledgments}
\else
  \section*{Acknowledgment}
\fi
The authors were supported by the MSMT LL1303 ERC-CZ grant. We would also like to thank Karel Lenc for insightful discussions.



\bibliographystyle{IEEEtran}
\bibliography{egbib}

%

\begin{IEEEbiography}[{\includegraphics[width=1in,height=1.25in,clip,keepaspectratio]{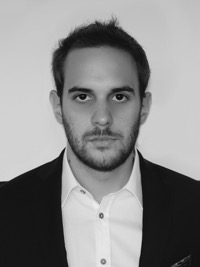}}]{Filip Radenovi{\'c}}
obtained his Master's degree from the Faculty of Electrical Engineering, University of Montenegro in 2013. Currently, he is a PhD candidate at the Visual Recognition Group, which is a part of the Department of Cybernetics, Faculty of Electrical Engineering, Czech Technical University in Prague. His research interests are mainly large-scale image retrieval problems.
\end{IEEEbiography}

\begin{IEEEbiography}[{\includegraphics[width=1in,height=1.25in,clip,keepaspectratio]{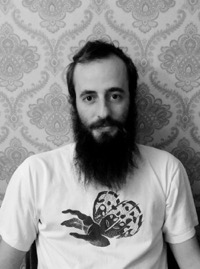}}]{Giorgos Tolias}
obtained his PhD from NTU of Athens and then moved to Inria Rennes  as a post-doctoral researcher. Currently, he is a post-doctoral researcher at the Visual Recognition Group of CTU in Prague. He enjoys working on large-scale visual recognition problems.
\end{IEEEbiography}


\begin{IEEEbiography}[{\includegraphics[width=1in,height=1.25in,clip,keepaspectratio]{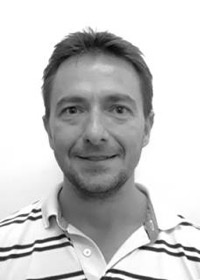}}]{Ond{\v r}ej Chum}
is leading a team within the Visual Recognition Group at the Department of Cybernetics, Faculty of Electrical Engineering, Czech Technical University in Prague. He received the MSc degree in computer science from Charles University, Prague, in 2001 and the PhD degree from the Czech Technical University in Prague, in 2005. From 2006 to 2007, he was a postdoctoral researcher at the Visual Geometry Group, University of Oxford, United Kingdom. His research interests include large-scale image and particular object retrieval, object recognition, and robust estimation of geometric models. He is a member of Image and Vision Computing editorial board, and he has served in various roles at major international conferences. He co-organizes Computer Vision and Sports Summers School in Prague. He was the recipient of the Best Paper Prize at the British Machine Vision Conference in 2002. He was awarded the 2012 Outstanding Young Researcher in Image and Vision Computing runner up for researchers within seven years of their PhD. In 2017, he was the recipient of the Longuet-Higgins Prize.
\end{IEEEbiography}





\end{document}